\pdfoutput=1

\documentclass[11pt]{article}

\usepackage[final]{acl}

\usepackage{times}
\usepackage{latexsym}
\usepackage[normalem]{ulem} 
\usepackage[T1]{fontenc}

\usepackage[utf8]{inputenc}

\usepackage{microtype}

\usepackage{inconsolata}

\usepackage{graphicx}
\usepackage{amsmath}
\usepackage{adjustbox}
\usepackage{multirow}
\usepackage{booktabs}
\usepackage{bbding}
\usepackage{array}
\usepackage{url}

\title{UniICL: An Efficient Unified Framework Unifying Compression, Selection, and Generation}

\author{Jun Gao$^1$, Qi Lv$^2$, Zili Wang$^4$, Tianxiang Wu$^1$,Ziqiang Cao$^{1\thanks{Corresponding Author}}$, Wenjie Li$^3$ \\
  School of Computer Science and Technology, Soochow University$^1$ \\
Harbin Institute of Technology (Shenzhen)$^2$ \\
  Hong Kong Polytechnic University$^3$
  Stepfun$^4$\\
\hypersetup{urlcolor=black}
{\normalsize \href{mailto:jgao1106@stu.suda.edu.cn}{jgao1106@stu.suda.edu.cn}, 
{\normalsize\href{mailto:zqcao@suda.edu.cn}{zqcao@suda.edu.cn}}
}}

\begin{document}
\maketitle



\begin{abstract}
In-context learning (ICL) enhances the reasoning abilities of Large Language Models (LLMs) by prepending a few demonstrations. 
It motivates researchers to introduce more examples to provide additional contextual information for the generation.
However, existing methods show a significant limitation due to the problem of excessive growth in context length, which causes a large hardware burden.
In addition, shallow-relevant examples selected by off-the-shelf tools hinder LLMs from capturing useful contextual information for generation.
In this paper, we propose \textbf{UniICL}, a novel \textbf{Uni}fied \textbf{ICL} framework that unifies demonstration compression, demonstration selection, and final response generation.
Furthermore, to boost inference efficiency, we design a tailored compression strategy that allows UniICL to cache compression results into \textbf{Demonstration Bank} (\textbf{DB}), which avoids repeated compression of the same demonstration.
Extensive out-of-domain evaluations prove the advantages of UniICL in both effectiveness and efficiency.

\end{abstract}

\begin{figure}[t]
    \centering
    \includegraphics[width=\columnwidth]{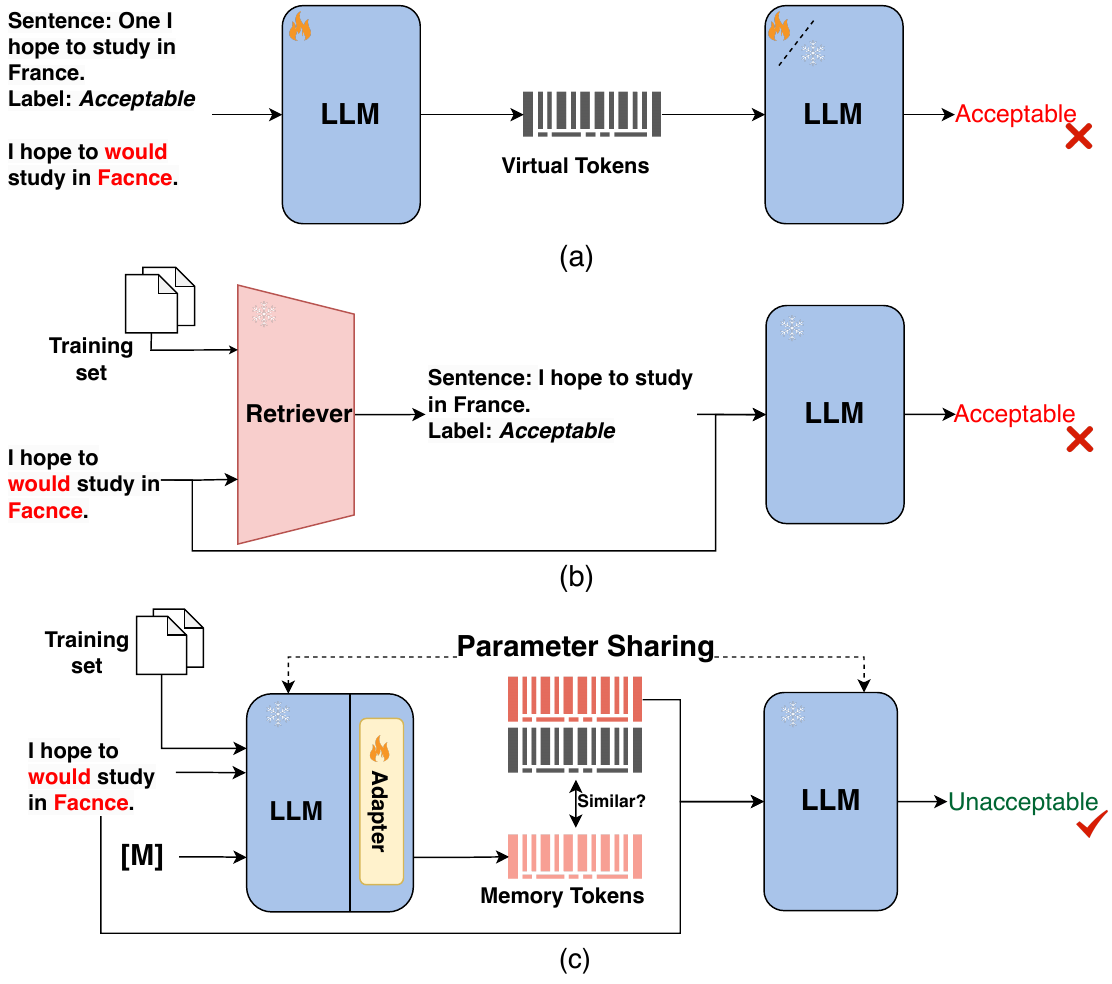}
    \caption{(a) Prompt compression methods that indiscriminately compress both demonstrations and queries.(b) Retrieval-based demonstration selection methods select lexical demonstrations. (c) UniICL discriminately compresses demonstrations and performs selection upon the compression results.}
    \label{fig:demo}
\end{figure}

\section{Introduction}
In-context learning (ICL)~\cite{brown2020language,xie2021explanation,wang2023label} to enhance the reasoning ability of Large Language Models (LLMs) with a few demonstrations prepended~\cite{wang2023chatgpt,yang2023exploring,wei2023zero,wang12023chatgpt,min2022noisy}.
Inspired by its outstanding performance, researchers explored applying ICL on many tasks such as text summarization~\cite{wang2023chatgpt,yang2023exploring,gao2024guiding}, sentiment classification, and linguistic acceptability~\cite{min2022noisy,wang2019glue}.
However, two challenges hinder the impact of ICL currently: (1) concatenated demonstrations directly surge the input length, causing a large hardware burden; (2) the prepended demonstrations are randomly sampled or selected via off-the-shelf tools which tend to provide shallow relevant demonstrations, hindering LLMs from capturing useful contextual information for generation.
Existing work tackles the two challenges separately.

To alleviate input length surge, on the one hand, many efforts are made in modifying model architecture to accommodate longer contexts~\cite{zheng2022linear,wu2022memorizing,ding2023longnet,bulatov2023scaling}.
These methods usually require training models from scratch, and models with a million context windows still struggle to overcome performance degradation~\cite{liu2024lost}.
On the other hand, recent studies attempt to shorten inputs through prompt compression~\cite{wingate2022prompt,mu2023learning,jiang2023llmlingua,ge2023context,gao2024selfcp}.
However, these compression methods are not applicable to ICL because they indiscriminately compress both demonstrations and queries into virtual tokens.
For instance, as illustrated in Fig.$\,$\ref{fig:demo}(a), the task entails justifying whether the query is grammatically acceptable.
The latter generator makes responses only according to virtual tokens generated by the compressor, resulting in a \textcolor{red}{wrong} answer\footnote{I hope to \sout{would} study in \sout{Facnce} (France)}.
More importantly, current compression methods are costly to train~\cite{wingate2022prompt,mu2023learning,jiang2023llmlingua}, and compressors are either limited to compressing within the original model's allowed input length~\cite{mu2023learning,jiang2023llmlingua,ge2023context} or bringing significant inference latency~\cite{wingate2022prompt}.

Retrieval-based In-context Example Selection (RICES) methods~\cite {alayrac2022flamingo} integrate an off-the-shelf pre-training model to select demonstrations similar to the queries at a shallow level.
These demonstrations usually contain redundant information and bring minimal benefits for the final generation~\cite{liu2021makes,ram2023context,wang2024large}.
Existing work attempts to train the retrieval model and the generator in an end-to-end manner, which has shown better performance in in-domain datasets~\cite{wang2023learning,qiao2024dntextspotter}.
However, this approach still performs poorly in out-of-domain datasets.
For instance, as shown in Fig.~\ref{fig:demo}(b), the retriever selects an example lexically similar to queries but has \textit{contrasting} labels.
Then, the LLM is misled and responds with a \textcolor{red}{wrong} answer.

In light of challenges in ICL, we turn to leverage the inherent understanding ability of LLMs developed during pre-training.
We accordingly propose a \textbf{Uni}fied \textbf{ICL} (\textbf{UniICL}) framework, which unifies demonstration compression, demonstration selection, and response generation.
As shown in Fig.~\ref{fig:demo}(c), for lightweight training, in UniICL, both the compressor and generator are initialized from the same LLM and kept frozen.
An adapter is introduced to align the compressor with the generator, and \textbf{[M]} is a learnable embedding called \textbf{Memory Slot} which is attached behind demonstrations for compression.
Therefore, UniICL only contains \textbf{17M} trainable parameters.
The LLM compressor first compresses each demonstration from the training set and queries into \textbf{Memory Tokens} independently on top of Memory Slots.
Then, UniICL selects $n$ most relevant demonstrations based on the similarity of Memory Tokens between queries and demonstrations.
Finally, Memory Tokens of selected demonstrations are concatenated to formulate a global in-context sequence, together with queries fed into the generator for response generation.
Due to independent compression, the compressor gets rid of the input window limitation of original LLMs as the number of demonstrations increases.
In addition to improvements in window limitation, the tailored compression strategy further makes improvements to ICL efficiency.
Specifically, UniICL caches Memory Tokens of different demonstrations to configure the \textbf{Demonstration Bank} (\textbf{DB}) for future reusing as shown in Fig.~\ref{fig:db}.
Therefore, repeated compression of the same demonstration is not necessary, which significantly boosts model efficiency in Fig.~\ref{fig:efficency}.
Extensive out-of-domain evaluation indicates UniICL achieves substantial improvements compared with other baselines.
Our main contributions are as follows:

\begin{figure}[t]
    \centering
    \includegraphics[width=0.48\textwidth]{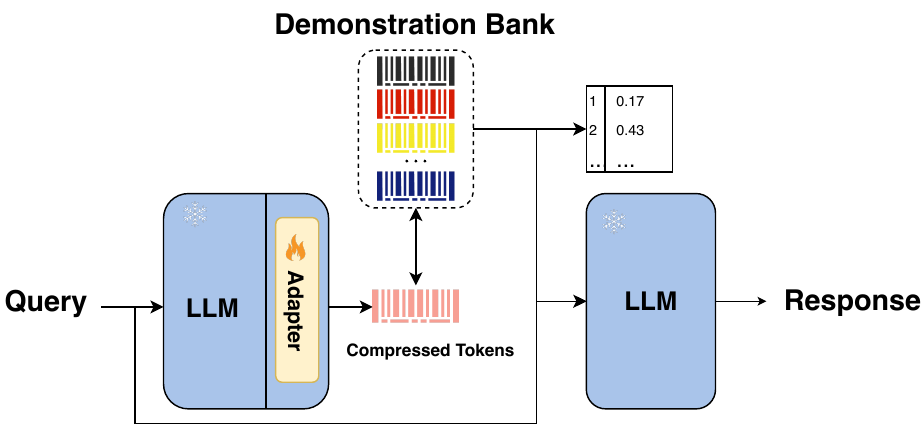}
    \caption{The workflow of Demonstration Bank.}
    \label{fig:db}
\end{figure}
\begin{itemize}
    \item To our knowledge, we are the first to propose a unified ICL framework with 17M trainable parameters.
    
    \item UniICL proposes configuring the Demonstration Bank to avoid repeated compression for the same demonstration, which significantly boosts ICL efficiency.

    \item Different from the indiscriminate compression of previous studies, UniICL proposes a tailored compression strategy for ICL, achieving substantial improvements compared with other baselines. 
\end{itemize}

\section{Related Work}
\subsection{Soft Prompt Compression}
Recently, researchers attempted to utilize soft prompts to convert actual tokens to dense-information virtual tokens.
Mostly from a distillation perspective, \citet{wingate2022prompt} aligned the teacher model and the student model, where the teacher model accepted the actual task instruction while the student model fed the soft prompt.
The main drawback of this approach was the lack of generalization that necessitated training for each lexically different instruction.
To tackle the generalization problem, \citet{mu2023learning} proposed to learn a Llama-7b to compress instructions to virtual tokens, but only compressing instructions was not powerful enough since the demonstrations were much longer in practice.
To compress longer prompts, \citet{chevalier2023adapting} proposed AutoCompressor to recurrently generate compressed virtual tokens based on a fine-tuned Llama \cite{zhang2022opt}.
However, AutoCompressor broke the independence of demonstrations, and the recurrent compression increased inference latency.
\citet{ge2023context} proposed ICAE that employed a LoRA-adopted Llama-7b \cite{touvron2023llama} to compress the processed demonstrations to compact virtual tokens, while ICAE still struggled to overcome quite long inputs.

\subsection{Extractive Compression}
Apart from employing soft prompts, researchers also endeavored to shorten prompts by extracting informative tokens from the original ones \cite{li2023unlocking,jiang2023llmlingua}, namely, token pruning \cite{kim2022learned} or token merging \cite{bolya2022token}.
Recent works like LLMLingua \cite{jiang2023llmlingua} and Selective Context \cite{li2023unlocking} shared similarities but diverged on whether to eliminate tokens with high or low Perplexity (PPL).
LLMLingua emphasized tokens with high PPL, attributing them as more influential, resulting in achieving outstanding performance.
As mentioned in their paper, extractive compression methods encountered Out-of-Distribution (OOD) issues between the extractor and the target LLM.
To reconcile this, they fine-tuned Alpaca-7b \cite{taori2023stanford} using the Alpaca dataset \cite{taori2023stanford} to perform the alignment.

\section{Methodology}
Previous compression methods are not tailored for ICL, and they are either bound by serious inference latency or poor performance, as demonstrated in Appendix~\ref{sec:comparsion}.
We propose UniICL, a unified ICL framework that unifies demonstration compression, demonstration selection, and response generation. 
As for the selection of the underlying LLM, previous work has proved that the Decoder-only model performs better than the Encoder-Decoder model in prompt compression \cite{mu2023learning}.
We follow this conclusion and adopt Vicuna-7B~\cite{zheng2023judging} as the underlying backbone in UniICL.

\begin{figure}
\centering
\includegraphics[width=.75\columnwidth]{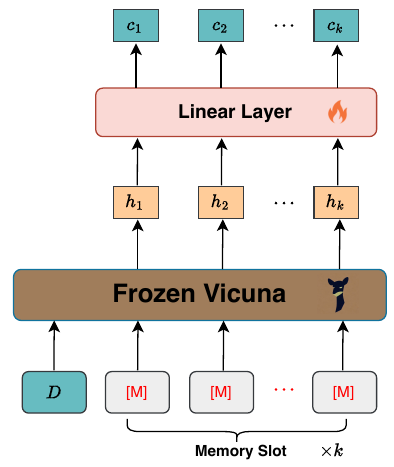}
\caption{Demonstration compression. $k$ Memory Slots are attached behind each demonstration.}
\label{fig:compress}
\end{figure}

\subsection{Demonstration Compression}
UniICL introduces Memory Slots $\textbf{\text{[M]}}\in \mathcal{R}^d$, a learnable $d$-dimension embedding initialized from a rarely used embedding of the target LLM.
UniICL activates the Memory Slots to extract information from demonstrations in the forward propagation $f_\theta(\cdot)$ of frozen Vicuna, as illustrated in Fig.~\ref{fig:compress}.
We first attach $k$ Memory Slots $M=k\times\text{[M]}$ behind each demonstration $D_i$, formatting modified prompt fed to the Vicuna.
Then, frozen Vicuna infers the modified prompts and outputs the last hidden states $H^i=(h_1, h_2, ..., h_k)$ on top of the $k$ Memory Slots:
\begin{equation}
    H^i = f_\theta(D_i^{L_i\times d} \oplus M^{k\times d}),
    \label{equ:enc}
\end{equation}where $L_i$ is the $i-$th demonstration length, $d$ is the embedding dimension and $\oplus$ means token-level concatenation.
Due to the attention mechanism, $H^i$ is compelled to attend to the preceding actual tokens.
Then, UniICL applies a linear layer as the adapter for efficiency to convert $H^i$ to Memory Tokens $C^i=(c^i_1, c^i_2, ..., c^i_k)$, performing alignment between the compressor and the generator\footnote{Linear layer is enough for UniICL as features have interacted with each other during compression.}:
\begin{equation}
c^i_j = W_p^{d\times d}\cdot h^i_j,
\label{equ:proj}
\end{equation}
where $W_p$ is the parameters of the projection layer.

\begin{figure}
\centering
\includegraphics[width=.75\columnwidth]{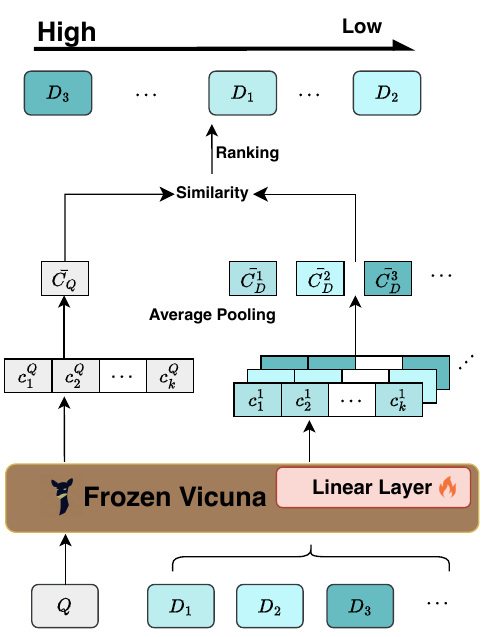}
\caption{Demonstrations selection.}

\label{fig:rerank}
\end{figure}
\subsection{Demonstration Selection}
Memory Tokens $C^i$ naturally summarize the demonstrations in latent space, and UniICL performs demonstration selection based on the similarity between queries and demonstrations as shown in Fig.~\ref{fig:rerank}.
Specifically, given a query $Q$ and its candidate demonstrations $(D_1, D_2, ..., D_n)$, UniICL obtains their representations used for selection by average pooling $C_{\{Q,D\}}$:
\begin{equation}
    \Bar{C^i}_{\{Q,D\}} = \frac{1}{k}\sum_{j=1}^k c_j.
    \label{equ:average}
\end{equation}
We define the $i$-th demonstration saliency score $S_i$ as the cosine similarity between $\bar{C_Q}$ and $\bar{D_i}$:
\begin{equation}
    S_i = \mathrm{cosine\_similarity}(\Bar{C_Q}, \Bar{C}_D^i).
\end{equation}

\subsection{Generation}
\begin{figure}[t]
\centering
\includegraphics[width=.75\columnwidth]{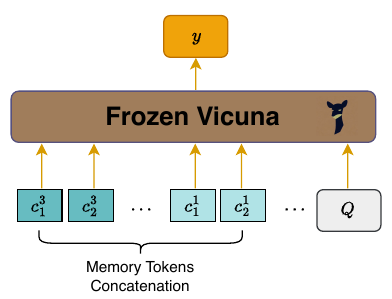}
\caption{In-context generation. The Memory Tokens from different demonstrations are concatenated horizontally at the input end of Vicuna.}
\label{fig:iclgeneration}
\end{figure}

We employ the frozen Vicuna again to generate responses with the guidance of concatenated Memory Tokens and queries, as illustrated in Fig.~\ref{fig:iclgeneration}.
For $m$-shot in-context learning, we obtain $m$ spans of Memory Tokens after demonstration compression and selection, denoted as $C^1$ to $C^m$.
Then, we horizontally concatenate them, keeping their relative position unmodified.
Finally, the concatenated Memory Tokens together with actual queries are fed into Vicuna, performing auto-regressive generation $g_\theta$ as normal:
\begin{equation}
    y_i = g_\theta(C^1, ..., C^m; Q; y_{<i}).
\end{equation}
Except for the generative manner, Memory Tokens apply close-ended evaluation for understanding tasks as normal through measuring the perplexity of candidate choices~\footnote{\url{https://huggingface.co/docs/transformers/perplexity}}.

\subsection{Training}
The trainable parameters in UniICL are merely 17M originating from the projection layer $W_p$ and the introduced Memory Slot [M].
The linear layer is optimized with the language modeling objective $\mathcal{L}_{lm}$ of Vicuna to learn a base compression model.
Then InfoNCE~\cite{he2020momentum} joint with language modeling objective are used to augment the demonstration selection ability of the base compression model:
\begin{equation}
    \mathcal{L}=\mathcal{L}_{lm} + \mathcal{L}_{ctr}.
    \label{equ:ctr}
\end{equation}
Specifically, we slice the source input of each training instance into two parts and randomly compress one.
The compressed part is denoted as $x_c$ and the uncompressed part is denoted as $x_u$.
Afterward, we attach the Memory Slot sequence $M$ behind $x_c$ and get Memory Tokens $C$ on top of the Memory Slots, as described in Eq.~\ref{equ:enc} and Eq.~\ref{equ:proj}.
Therefore, the language modeling loss $\mathcal{L}_{lm}$ is obtained as:
\begin{equation}
    \mathcal{L}_{lm} = -\frac{1}{|y|}\sum_{t=0}logP(y_t|x_u; C; y_{<t}),
    \label{eq:lm}
\end{equation}
where $y$ is the reference label of the current training instance.
Additionally, to approach the large-shot settings without significant truncation, we introduce concatenation compression.
When $x_c$ exceeds the window limitation for compression, UniICL further divides $x_c$ into acceptable ranges and compresses them independently to get local Memory Tokens.
Then, these Memory Tokens from different segments will be concatenated to formulate global virtual tokens to replace $x_c$, applying Eq.~\ref{eq:lm} to optimize models as well.

\begin{figure}
\centering
\includegraphics[width=.75\columnwidth]{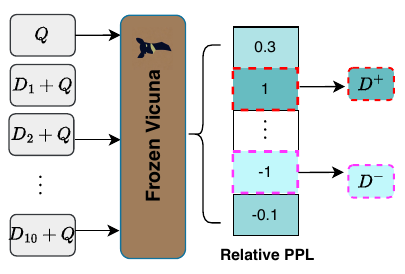}
\caption{Contrastive examples mining pipeline. Finds demonstrations benefit/hinder the final generation according to the PPL.}
\label{fig:negatives}
\end{figure}

We obtained a base compression model that has learned to compress and understand concatenated Memory Tokens after the first-phase training mentioned.
Subsequently, we utilize contrastive learning for selection augmentation and mine positives and negatives as illustrated in Fig.~\ref{fig:negatives}.
Specifically, given each training instance $Q$ and $n$ candidate demonstrations $(D_1, D_2, ..., D_n)$ from two non-crossing training subsets, we employ Vicuna to calculate the PPL concerning the golden label of $Q$, denoted as $ppl^Q$ to find useful demonstrations for generation.
Then, we provide the $i$-th demonstration and calculate PPL concerning the golden label of $Q$, denoted as $(ppl^D_i, i \in [1, n])$.
We count $ppl^Q$ as the baseline and calculate candidate relative PPL gains:
\begin{equation}
   \widetilde{ppl}^D_i=ppl^Q - ppl^D_i, i \in [1, n].
\end{equation}
After finding demonstrations $D^+$ ($D^-$) that furthest reduces (increases) $ppl^Q$, we obtain their representation $C_D^+$ $(C_D^-)$ as processed in Eq.~\ref{equ:average}.
The contrastive loss $\mathcal{L}_{ctr}$ can be formulated as:
\begin{equation}
    \mathcal{L}_{ctr}=\frac{\exp(cos(C_Q, C_D^+))}{\exp(cos(C_Q, C_D^+)) + \exp(cos(C_Q, C_D^-))}.
\end{equation}
In particular, if all relative PPL gains are less than 0, namely none of the candidate demonstrations help guide Vicuna to generate the golden label, we will apply the other set of candidates.

\section{Experiment}

\subsection{Baselines}
Unmodified Vicuna-7b serves as the fundamental baseline fed with actual demonstrations.
AutoCompressor compresses prompts into 50 virtual tokens in different rounds recurrently.
Previous compressed virtual tokens are put at the beginning of the current segment.
Finally, virtual tokens of different compression rounds are concatenated for generation.
We employ their Llama2-7b version for comparison.
LLMLingua is a coarse-to-fine demonstration pruning method based on dropping uninformative words.
We employ their released 7b version, of which the compressor is a fine-tuned Llama2.
For a meaningful comparison, we replace target LLMs of LLMLingua (GPT-3.5-Turbo or Claude-v1.3) with the Vicuna-7b.
ICAE compresses demonstrations into 128 virtual tokens via a LoRA-adapted Llama2-7b.
Additionally, since selection augmentation is involved in the training of UniICL, we utilize the popular Sentence-BERT (S-BERT)~\cite{reimers2019sentence} as the dense retriever to construct an ICL pipeline for the above methods, serving as simple but effective selection-based baselines.

\subsection{Settings}
\begin{table}
\centering
\begin{adjustbox}{width={0.45\textwidth},totalheight={\textheight},keepaspectratio}
\begin{tabular}{l|ccc}
\bottomrule
\multirow{2}{*}{Dataset}  &\multicolumn{3}{c}{\# words} \\
&(96,512]&(512,1024]&(1024,1536]\\
\hline
XSum~\cite{Narayan2018DontGM} &-&10,000&4,697\\
\hline 
CICERO~\cite{ghosal2022cicero}&10,000&-&-\\
\hline
SUPER-NI~\cite{wang2022super} & - & 10,000& 7,000\\
\hline
XSum (Ctr) & \multicolumn{3}{c}{5,000}\\

\toprule
\end{tabular}
\end{adjustbox}
\caption{The composition training set of UniICL. (m,n] represents the range of the number of words in each instance. XSum (Ctr) is used for the second-phase training in Eq.~\ref{equ:ctr}.}
\label{tab:dataset}
\end{table}

\begin{table}
  \centering
    \begin{adjustbox}{width={0.48\textwidth},totalheight={\textheight},keepaspectratio}
    \begin{tabular}{lcccc}
    \bottomrule
    Dataset    & In-Domain &\# Test & \# Demonstrations \\
    \hline
    MS MARCO-dev  & \XSolidBrush & 6,980 & -\\
    \hline
    XSum    & \Checkmark & 1,500 & 204,045/20\\
    \hline
    Arxiv    & \XSolidBrush & 1,500 & 203,037/20\\
    \hline
    CoLA-dev  &  \XSolidBrush & 1,041 &67,349/20\\
    \hline
    SST-2-dev    &  \XSolidBrush & 872 & 8,551/20 \\
    \hline
    IMDb   & \XSolidBrush & 1,500 & 25,000/20\\
    \hline
    MMLU   & \XSolidBrush & 13,985 & 25,000/20\\
    \toprule
    \end{tabular}%
\end{adjustbox}
  \caption{The details of the involved evaluation datasets. -dev represents employing the development set due to their test sets are inaccessible. \# Demonstrations represent the number of demonstrations to be selected in \textbf{high}/{low}-resource ICL settings.}
  \label{tab:settings}%
\end{table}%

\begin{figure}[t]
\centering
\includegraphics[width=0.95\columnwidth]{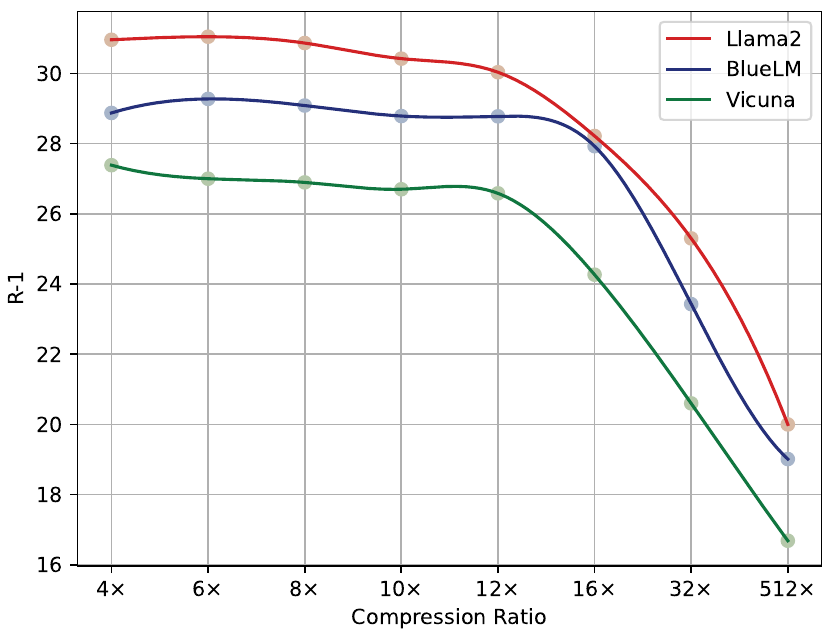}
\caption{The compression ratio sensitivity analysis of 
\textcolor[RGB]{210,34,37}{Llama2}
, \textcolor[RGB]{37,48,122}{BlueLM}, and \textcolor[RGB]{16,118,63}{Vicuna}.}
\label{fig:ratio}
\end{figure}

We construct the training set by mixing up XSum, CICERO, and SUPER-NI according to their length as shown in Tab.~\ref{tab:dataset} and evaluate UniICL on extensive out-of-domain datasets as listed in Tab.~\ref{tab:settings}, with more details reported in Appendix~\ref{sec:DM}. 
Considering computation efficiency, we set the max allowed input length limit to 512 for both compression and generation for both training and inference.
For a fair comparison, we set the allowed window of baselines to 512, and the compression ratio of default UniICL and baselines is set to 12, which is determined by the validation in Fig.~\ref{fig:ratio}.
We fix the learning rate to 8e-5 and use Adam as the optimizer, and the effective batch size is 32 (8 GPUs data parallelism and 4 steps gradient accumulation).
We train 10 epochs and 2 epochs respectively for the first- and second-phase training.
The best checkpoints are selected according to their performance on in-domain validation sets.
Additionally, we conducted all experiments on 8*NVIDIA A5000 24G GPUs based on BFloat 16 data type, and we set the evaluated shots to 8 for understanding tasks and 5 for generative tasks for illustration, because of marginal ICL gains and memory costs.

We apply S-BERT to pre-rank and output the top 10 similar candidates from training sets according to each inference input for all baselines.
UniICL is employed to perform selection among them in practice due to computational efficiency for high-resource ICL.
On the contrary, the low-resource ICL setting utilizes the randomly sampled 20 candidate demonstrations for all inference inputs, while UniICL performs selection as normal.

To verify the universality, we further build UniICL on BlueLM-7B~\cite{2023bluelm} and Llama2-7B~\cite{touvron2023llama}.
Results of BlueLM and Llama2 will be reported in Appendix~\ref{sec:bluelm} and Appendix~\ref{sec:llama2}.
\subsection{Results}
\label{sec:results}
\begin{table*}[t]
\centering
\begin{adjustbox}{width={0.99\textwidth},totalheight={\textheight},keepaspectratio}
\begin{tabular}{l|c|ccc|ccc|ccc}
\bottomrule
\multirow{2}{*}{Model} & \multirow{2}{*}{\#-shots}  &CoLA-dev  &SST-2-dev & IMDb &\multicolumn{3}{c|}{Arxiv} &\multicolumn{3}{c}{XSum} \\
 &   &\multicolumn{3}{c|}{Acc.}  & R-1 & R-2 & R-L & R-1 & R-2 & R-L \\

\hline
\multirow{4}{*}{Vicuna}&0-shot &56.2   &91.7   &92.6 & 34.3 & 9.1 & 27.4 &19.9& 5.0 & 13.4\\

 &1-shot &58.2 (57.4)&90.7 (90.8) &91.9 (91.0) & 34.8 (34.4)& 9.3 (9.1)& 27.9 (27.5) & 21.5 (21.2) & 5.9 (5.8) & 14.7 (14.5) \\
& 2-shot &62.1 (59.8)&92.1 (91.3) &91.7 (91.7) & -& -&-& -& -&- \\
 & 5-shot &62.3 (61.9)&93.0 (91.9)&94.1 (92.5) & -& -&- & -& -&-\\
\hline
\multirow{3}{*}{AutoCompressor}  &1-shot &42.1 (40.9) &85.7 (84.2)&95.0 (95.1) & 27.0 (26.4) & 8.4 (8.2) & 26.1 (25.8) & 21.3 (20.3) & 6.5 (6.3) & 13.7 (13.7)\\
& 2-shot &58.8 (56.3) & 88.0 (86.4)& 95.0 (94.6) & 27.1 (26.2) & 8.6 (7.9) & 26.4 (25.4) & 21.9 (21.4) & 6.6 (6.4)& 14.5 (14.1)  \\
 & 5-shot &59.1 (58.8) & 91.3 (89.1)& 94.7 (94.8) & 34.5 (33.7)& 9.4 (9.1)&28.7 (27.9) & 22.4 (21.7) & 6.9 (6.7) & 14.8 (14.3)\\
\hline

\multirow{3}{*}{LLMLingua}  &1-shot & 55.5 (55.0) & 89.7 (89.6)& 91.0 (89.9) & 33.3 (33.1) & 8.9 (8.7) & 27.4 (27.1) & 20.5 (19.7) & 5.4 (5.2) & 14.5 (14.4)\\
& 2-shot & 56.7 (55.7) & 90.7 (90.2) & 91.3 (91.0) & 32.9 (32.0) & 8.2 (8.1) & 26.9 (25.9) & 20.3 (20.0) & 5.2 (5.1) & 14.3 (14.1)  \\
           
 & 5-shot & 57.2 (56.9) & 90.6 (90.2)& 90.9 (91.2) & 30.1 (29.7)& 7.9 (7.4)& 25.3 (24.6) & 19.7 (18.6) & 4.9 (4.9) & 14.1 (14.3) \\
\hline

\multirow{3}{*}{ICAE}  &1-shot & 30.9 (30.9) & 61.0 (60.1)& 85.7 (83.3) & 26.8 (24.6) & 8.2 (7.1) & 24.7 (22.9) & 23.5 (21.9) & 8.5 (7.8) & 20.9 (20.3)\\
& 2-shot &30.9 (30.9)& 49.0 (52.8) & 85.9 (85.9) & 27.2 (25.5) & 8.4 (7.6) & 25.9 (24.3) & 24.4 (23.2) & 8.9 (8.4) & 21.3 (20.8) \\
 & 5-shot &30.9 (30.9) & 54.2 (51.0) & 85.7 (85.9) & 28.3 (26.9) & 8.7 (7.7)& 26.6 (25.8) & 25.3 (24.9) & 9.2 (8.8) & 22.5 (21.6)\\
\hline

\multirow{3}{*}{UniICL} &1-shot &58.7 (58.0)&92.9 (91.7)&94.3 (92.3) & 35.5 (34.7) & 10.5 (10.2) & 28.7 (27.9) &27.7 (25.5) &10.2 (9.1) & 21.2 (20.0)\\
 & 2-shot &62.4 (61.0)&92.4 (91.6) & 94.9 (93.3) & 36.1 (35.2) & \textbf{10.8} (10.4) & 29.4 (28.2) &29.4 (26.8) &11.0 (9.8) & 22.3 (20.9) \\
& 5-shot &62.6 (61.8)&93.1 (92.3) &94.5 (94.0) & 35.8 (35.4) & 10.6 (10.2)& 29.5 (28.1) & 30.7 (27.6) & 11.3 (10.1) & 22.8 (21.4)\\
\hline

\multirow{4}{*}{UniICL$^\spadesuit$} &1-shot &59.1 (58.7) & 93.0 (91.9) & 94.5 (91.6) & 34.8 (34.7) & 10.4 (10.3) & 28.1 (27.8) &29.1 (26.2) &10.8 (9.4)  & 22.2 (20.7)\\
&2-shot &62.6 (61.2) & 94.0 (93.0) & 94.9 (92.3)  & 34.6 (34.3) & 10.6 (10.4)& 28.5 (28.3) &30.3 (28.9)& 11.3 (10.5) & 22.9 (21.7)\\
& 5-shot  &63.3 (61.5)&\textbf{94.7} (92.8)  & 95.0 (93.8) & 35.6 (35.3) & 11.0 (10.8)& 29.1 (27.7) & 31.1 (30.0) & 11.7 (11.2)&  23.5 (22.3)\\
& 8-shot  &63.8 (62.6)&\textbf{94.7} (93.1)  & 95.0 (94.2) & -& -&- & -& -&-\\

 \hline
 
\multirow{4}{*}{UniICL$^\spadesuit$ + $L_{ctr}$} &1-shot & 59.3 (58.9) & 93.2 (92.4) & 95.1 (92.8) & 35.6 (35.1) & 10.7 (10.5) & 28.9 (28.3) &30.0 (27.9) &11.3 (10.1) & 22.8 (21.5)\\
 &2-shot &62.4 ({62.0}) & 94.5 (92.8) & 94.8 (93.4) &{36.8} (\underline{35.3})& {10.8} ({10.6})&{29.6} (\underline{28.9}) &{30.8} ({29.2}) &{11.4} ({10.7}) &{23.0} ({21.9})\\
 
 &5-shot &{64.3} (61.8)&\textbf{94.7} ({93.4}) & \textbf{96.1} ({94.2}) &\textbf{37.1} ({34.9})& \textbf{11.3} (\underline{11.2})&\textbf{30.0} (\underline{29.3}) & \textbf{32.5} (\underline{30.6}) & \textbf{12.3} (\underline{11.8}) & \textbf{24.7} (\underline{23.8})\\
 
&8-shot &\textbf{64.7} (\underline{63.3})&\textbf{94.7} (\underline{94.1}) & {95.6} (\underline{95.0}) & -& -&-  & -& -&- \\

\toprule
\end{tabular}
\end{adjustbox}
\caption{The high- and low-ICL results on CoLA-dev, SST-2-dev, and IMDb. Results in (bracket) represent low-resource ICL.
$^\spadesuit{}$ represents the demonstrations selected by UniICL, and the others are selected by S-BERT.
+$L_{ctr}$ indicates the selection augmented UniICL (optimized with Eq.~\ref{equ:ctr}). \textbf{Bold} (\underline{underline}) represents the best performance on high- and low-resource ICL. R- indicates Rouge scores. All compression methods are evaluated with a compression ratio set to 12.}
\label{tab:understanding}
\end{table*}

We comprehensively evaluate the ICL performance of UniICL on the out-of-domain dataset CoLA, SST-2, and IMDb by close-ended evaluation and Arxiv by open-ended evaluation in Tab.~\ref{tab:understanding}.
The details of the involved evaluation datasets and metrics are reported in Tab.~\ref{tab:settings} and Appendix~\ref{sec:DM}.
Specifically, UniICL outperforms unmodified Vicuna-7b fed with actual candidate demonstrations, which indicates that Memory Tokens are more efficient and informative for guiding the target LLM.
Meanwhile, UniICL outperforms all the baselines by compressing the same demonstrations pre-ranked by S-BERT.
Additionally, UniICL achieves further performance gains after selecting demonstrations via itself (UniICL$^\spadesuit$).
The open-ended results highlight that Memory Tokens indeed capture semantic information for ICL generation, even though summarization demonstrations are much longer than understanding ones.
Regarding Arxiv, the original ICL is not helpful enough due to its extremely over-length document, leaving little room for demonstrations.
UniICL works as expected by compressing demonstrations into Memory Tokens and concatenating them, achieving +2.8 Rouge-1 gains in selection-augmented UniICL (+$\mathcal{L}_{ctr}$).
Additionally, according to the results of +$\mathcal{L}_{ctr}$, we find that the gains brought by selection augmentation become larger as the number of demonstrations increases.
We attribute this to the fact that UniICL selects more useful demonstrations for generation after the second-phase training.
The results of BlueLM are exhibited in Appendix~\ref{sec:bluelm}.
Except for understanding and generative tasks, we further evaluate UniICL on MMLU in Tab.~\ref{tab:mmlu}.
UniICL achieves stable performance gains with more demonstrations introduced.
Additionally, considering ICAE and AutoCompressor are soft-prompt-based compression methods built on Llama2, we also build UniICL on Llama2 for ablation in Appendix~\ref{sec:llama2}.

\begin{table}
\centering
\begin{adjustbox}{width={0.8\columnwidth},totalheight={\textheight},keepaspectratio}
\begin{tabular}{l|cccc|c}
\bottomrule
{\#-Shots} & S & H & SS & O & Avg. \\ 
\hline
0-shot     & 36.9 & 53.2       & 53.7            & 50.7  & 48.6    \\
1-shot     & 38.6 & 55.3       & 54.6            & 52.4  & 50.2    \\

2-shot     & 39.2 & \textbf{55.8}       & \textbf{55.3}            & 53.1  & 50.9    \\

5-shot     & \textbf{40.1} & 55.6       & \textbf{55.3}            & \textbf{53.8}  & \textbf{51.2}    \\
\toprule
\end{tabular}
\end{adjustbox}
\caption{Performance of UniICL on MMLU benchmark. We reported the Accuracy at the category level. S represents \textbf{S}TEM, H represents \textbf{H}umanities, SS represents \textbf{S}ocial \textbf{S}cience, O represents \textbf{O}ther, and Avg indicates their average performance. }
\label{tab:mmlu}
\end{table}

\paragraph{Passage Ranking}
\begin{table}
\centering
\begin{adjustbox}{width={0.8\columnwidth},totalheight={\textheight},keepaspectratio}
\begin{tabular}{l|cc}
\bottomrule
Method & \# TP  & MRR@10 \\
\hline
BM25$^\dagger$ & - &18.5\\
Vicuna & - & 28.9\\
AutoCompressor& - & 29.3 \\
ICAE & - & 30.2\\
UniICL& - & \textbf{31.6}\\
\hline
SIMLM$^{\dagger\ddagger}$ & 110M & \underline{41.1}\\
UniICL$^\ddagger$ & 17M & {38.9}\\

\toprule
\end{tabular}
\end{adjustbox}
\caption{MRR@10 results on MS MARCO. Vicuna applies the last hidden states of [EOS] to represent sentences in latent space. Results citing from Liang~\cite{wang2022simlm} are denoted as $^\dagger$, and methods supervised trained on MS MARCO are represented as $^\ddagger$. \textbf{Bold} indicates the best zero-shot performance and \underline{Underline} is the best fine-tuned results. \# TP indicates the number of trainable parameters.}
\label{tab:passage}
\end{table}
Since the virtual tokens naturally summarize semantic information of preceding sequences, we evaluate UniICL on the out-of-domain MS MARCO dataset in Tab. \ref{tab:passage}.
UniICL significantly outperforms the sparse retrieval method BM25 algorithm and other compression methods.
Subsequently, we fine-tune the first-phase compression model of UniICL on the training set of MS MARCO.
UniICL achieves comparable performance with SIMLM~\cite{wang2022simlm}, which is specified in Information Retrieval (IR) and has more trainable parameters.

\section{Analysis}

\subsection{Compression Ratio}
\begin{table}
\centering
\begin{adjustbox}{width={0.49\textwidth},totalheight={\textheight},keepaspectratio}
\begin{tabular}{l|ccc|c}
\bottomrule
\multirow{2}{*}{\#-shots} & {CoLA} & {SST-2} & {IMDb} & {Arxiv} \\
& \multicolumn{3}{c|}{Acc.} & R-1 \\
\hline
{1-shot} & 58.5 (\textcolor{red}{-0.8}) & 91.4 (\textcolor{red}{-1.8}) & 92.6 (\textcolor{red}{-2.5}) & 34.8 (\textcolor{red}{-0.8}) \\ 
{2-shot} & 59.7 (\textcolor{red}{-2.7}) & 92.1 (\textcolor{red}{-2.4}) & 94.1 (\textcolor{red}{-0.7}) & 35.7 (\textcolor{red}{-1.1}) \\ 
{5-shot} & 62.4 (\textcolor{red}{-1.9}) & 93.1 (\textcolor{red}{-1.6}) & 94.8 (\textcolor{red}{-1.3}) & 36.6 (\textcolor{red}{-0.5}) \\ 
\toprule
\end{tabular}
\end{adjustbox}
\caption{Performance of UniICL on out-of-domain datasets, with a fixed compression ratio set to 12 during training.}
\label{tab:fixratio}
\end{table}

During training, the compression ratio is dynamically sampled from 2 to 16.
We mix up 2,000 instances from the in-domain validation set, 1,000 for XSum, and 1,000 for CICERO to select the compression ratio for UniICL in Fig.~\ref{fig:ratio}, with the backbone of Llama2, Vicuna, and BlueLM respectively.
Specifically, UniICL compresses the latter cut-off part while keeping the former ones uncompressed.
Therefore, we can measure the dense information quality of the same content with different compression ratios by ROUGE-1 since it is more sensitive to token-level differences.
The performance is relative smoothing when the compression ratio changes from $4\times$ to $12\times$.
However, when it comes to $16\times$, an obvious drop occurs.
In order to analyze this phenomenon more deeply, we provide a thorough analysis in Appendix~\ref{sec:visualization}.
Therefore, we set the compression ratio to 12 by default and apply this ratio to all experiments.
The $512\times$ compression ratio is equal to compressing anything to a single virtual token, due to the maximum allowed input length for compression being 512.

\begin{figure}
    \centering
    \includegraphics[width=\linewidth]{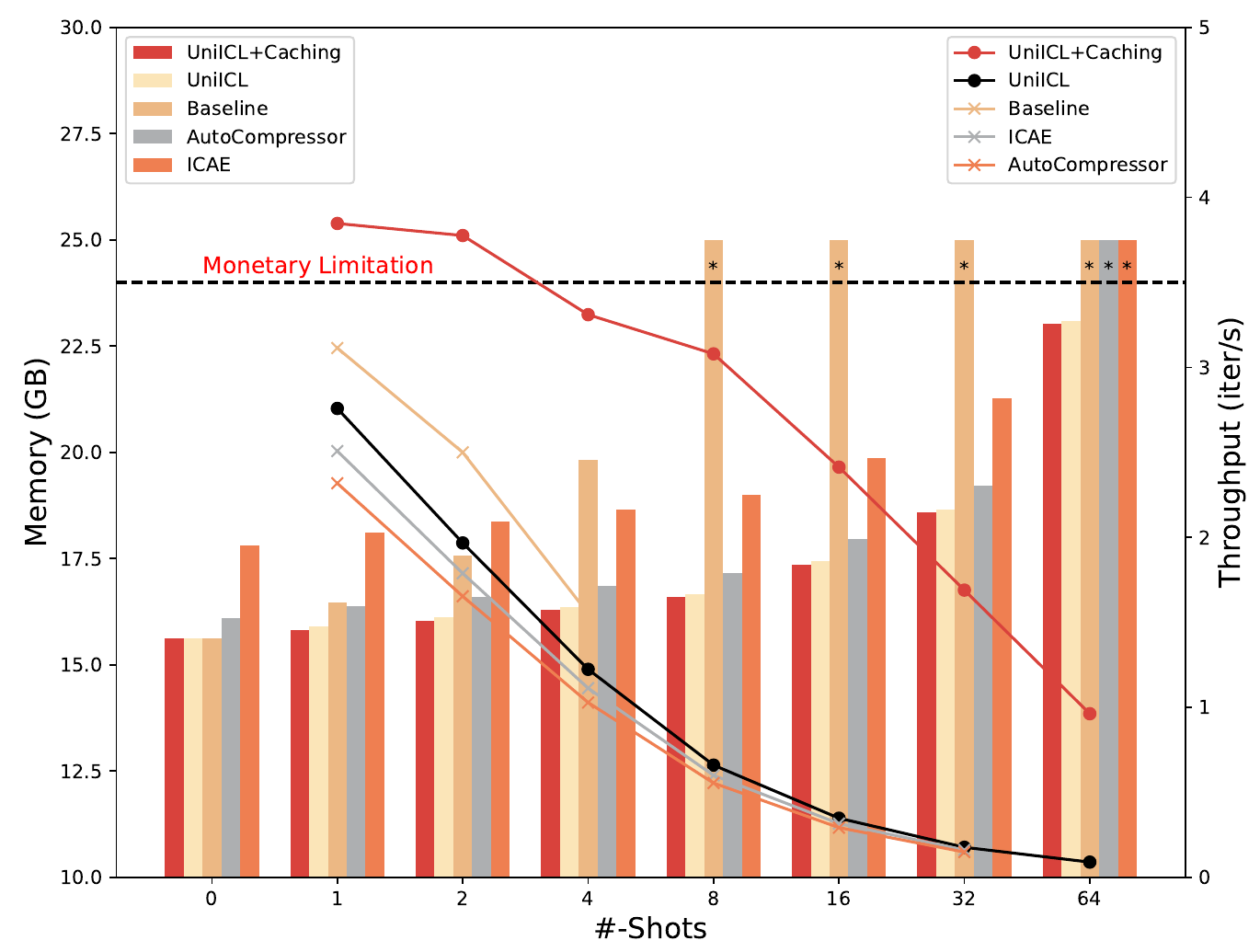}
    \caption{The efficiency comparison between UniICL and other compression methods in CoLA with the number of shots increasing from 0 to 64.
    Memory explodes are represented as *, corresponding to the break of the line chart. +Caching represents using DB.}
    \label{fig:efficency}
\end{figure}
To explore whether it could yield additional performance gains compared with dynamic ratios, in Tab.~\ref{tab:fixratio}, we re-train UniICL with the compression ratio fixed to 12 (Results of more fixed ratios are reported in Appendix~\ref{sec:fixedratio}.).
Results indicate that UniICL trained with fixed compression ratios underperforms in out-of-domain datasets as it exhibits over-fitting in in-domain sets as shown in Tab.~\ref{fig:fixratio}.

Furthermore, we analyze whether 12$\times$ is suitable for all out-of-domain datasets in Fig.~\ref{fig:winrate} in Appendix~\ref{sec:ratioselection}.
Results indicate that 12$\times$ outperforms other compression ratios in general across 4 out-of-domain datasets.
It also points out that lower ratios still work comparable for short demonstrations and higher ratios are suitable for long demonstrations to some extent.

\begin{table}
\centering
\begin{adjustbox}{width={0.45\textwidth},totalheight={\textheight},keepaspectratio}
\begin{tabular}{l>{\centering\arraybackslash}p{1.8cm} >{\centering\arraybackslash}p{1.8cm}>{\centering\arraybackslash}p{1.8cm}}
\bottomrule
Method & GPUHours & TFLOPs & TMACs\\
\hline
Vicuna & 1.5 & 86,20 & 4,309\\
Vicuna-1k & 1.9 &31,664 & 15,832  \\

UniICL& 1.6 & 22,437 & 11,218\\
\toprule
\end{tabular}
\end{adjustbox}
\caption{The computation efficiency of UniICL.}
\label{tab:efficiency}
\end{table}

\subsection{Efficiency Analysis}
In UniICL, we incorporate an additional 17M trainable parameters into the 7b backbone, accounting for an approximate increase of 0.24\%.
We evaluate the memory costs and inference latency of UniICL and other compression methods in Fig.~\ref{fig:efficency}.
With the help of the Demonstration Bank (DB), UniICL will eliminate the extra latency if the selected demonstrations have been compressed and cached (UniICL+Caching).
Despite this, parallel computation facilitates the compression process, resulting in minimal throughput degradation (UniICL and Baseline).
The unmodified 7B LLM causes a memory explosion for 8-shot settings, and other compression methods perform up to 32-shot, while UniICL successfully scales up to 64-shot within a 24GB CUDA allocation. 

Additionally, we demonstrate the inference computation and GPU hours in Tab.~\ref{tab:efficiency}, by using 1,024 random legal tokens as inputs and forcing models to generate 128 tokens.
Notably, UniICL (without DB) compresses the former half, and the latter half is fed into the generator directly, while Vicuna and Vicuna-1k are distinguished in window limitations.
Results indicate that minimal GPU hours increased due to the parallel computation of forward, although the extra compression of UniICL surges the computation.
Additionally, Vicuna, with a 1k window limitation, surges both GPU hours and TFLOPs because long input brings significant computation and latency in generation.

\section{Conclusion}
This paper proposes UniICL, a parameter-efficient ICL framework that unifies demonstration selection, demonstration compression, and final response generation via a frozen LLM, an adapter, and a learnable embedding.
Experimental results prove the advantages of UniICL in both efficiency and effectiveness.
Due to 12$\times$ demonstration compression, UniICL scales up the number of demonstrations from 4 to 64 within a 24 GB VRAM allocation.
Finally, to avoid repeated compression of the same demonstration, UniICL configures a Demonstration Bank (DB, which significantly boosts model efficiency.

\section{Limitations}
Our study, while proposing an efficient unified ICL framework for demonstration compression and selection, still has limitations.
Firstly, UniICL is limited to the realm of unmodified ICL, leaving other advanced LLM prompting methods, e.g., Retrieval Augment Generation (RAG) and Chain-of-Thought (CoT), unexplored.
Limited to the hardware, we deploy the underlying LLM at a scale of 7 billion parameters.
Larger-scale LLMs are welcome to enrich our findings in future studies.

\section{Acknowledgement}
I would like to express my sincere gratitude to all the authors and reviewers for their valuable contributions to this research.
This work was supported by the National Natural Science Foundation of China (NSFC 62106165) and the Project Funded by the Priority Academic Program Development of Jiangsu Higher Education Institutions, China.

\bibliography{custom}

\begin{thebibliography}{47}
\providecommand{\natexlab}[1]{#1}

\bibitem[{Alayrac et~al.(2022)Alayrac, Donahue, Luc, Miech, Barr, Hasson, Lenc, Mensch, Millican, Reynolds et~al.}]{alayrac2022flamingo}
Jean-Baptiste Alayrac, Jeff Donahue, Pauline Luc, Antoine Miech, Iain Barr, Yana Hasson, Karel Lenc, Arthur Mensch, Katherine Millican, Malcolm Reynolds, et~al. 2022.
\newblock Flamingo: a visual language model for few-shot learning.
\newblock \emph{Advances in neural information processing systems}, 35:23716--23736.

\bibitem[{Bolya et~al.(2022)Bolya, Fu, Dai, Zhang, Feichtenhofer, and Hoffman}]{bolya2022token}
Daniel Bolya, Cheng-Yang Fu, Xiaoliang Dai, Peizhao Zhang, Christoph Feichtenhofer, and Judy Hoffman. 2022.
\newblock Token merging: Your vit but faster.
\newblock \emph{arXiv preprint arXiv:2210.09461}.

\bibitem[{Brown et~al.(2020)Brown, Mann, Ryder, Subbiah, Kaplan, Dhariwal, Neelakantan, Shyam, Sastry, Askell et~al.}]{brown2020language}
Tom Brown, Benjamin Mann, Nick Ryder, Melanie Subbiah, Jared~D Kaplan, Prafulla Dhariwal, Arvind Neelakantan, Pranav Shyam, Girish Sastry, Amanda Askell, et~al. 2020.
\newblock Language models are few-shot learners.
\newblock \emph{Advances in neural information processing systems}, 33:1877--1901.

\bibitem[{Bulatov et~al.(2023)Bulatov, Kuratov, and Burtsev}]{bulatov2023scaling}
Aydar Bulatov, Yuri Kuratov, and Mikhail~S Burtsev. 2023.
\newblock Scaling transformer to 1m tokens and beyond with rmt.
\newblock \emph{arXiv preprint arXiv:2304.11062}.

\bibitem[{Chevalier et~al.(2023)Chevalier, Wettig, Ajith, and Chen}]{chevalier2023adapting}
Alexis Chevalier, Alexander Wettig, Anirudh Ajith, and Danqi Chen. 2023.
\newblock Adapting language models to compress contexts.
\newblock \emph{arXiv preprint arXiv:2305.14788}.

\bibitem[{Ding et~al.(2023)Ding, Ma, Dong, Zhang, Huang, Wang, and Wei}]{ding2023longnet}
Jiayu Ding, Shuming Ma, Li~Dong, Xingxing Zhang, Shaohan Huang, Wenhui Wang, and Furu Wei. 2023.
\newblock Longnet: Scaling transformers to 1,000,000,000 tokens.
\newblock \emph{arXiv preprint arXiv:2307.02486}.

\bibitem[{Gao et~al.(2024{\natexlab{a}})Gao, Cao, Huang, Qin, and Ai}]{gao2024guiding}
Jun Gao, Ziqiang Cao, Shaoyao Huang, Luozheng Qin, and Chunhui Ai. 2024{\natexlab{a}}.
\newblock Guiding chatgpt to generate salient domain summaries.
\newblock \emph{arXiv preprint arXiv:2406.01070}.

\bibitem[{Gao et~al.(2024{\natexlab{b}})Gao, Cao, and Li}]{gao2024selfcp}
Jun Gao, Ziqiang Cao, and Wenjie Li. 2024{\natexlab{b}}.
\newblock Selfcp: Compressing over-limit prompt via the frozen large language model itself.
\newblock \emph{Information Processing \& Management}, 61(6):103873.

\bibitem[{Ge et~al.(2023)Ge, Hu, Wang, Chen, and Wei}]{ge2023context}
Tao Ge, Jing Hu, Xun Wang, Si-Qing Chen, and Furu Wei. 2023.
\newblock In-context autoencoder for context compression in a large language model.
\newblock \emph{arXiv preprint arXiv:2307.06945}.

\bibitem[{Ghosal et~al.(2022)Ghosal, Shen, Majumder, Mihalcea, and Poria}]{ghosal2022cicero}
Deepanway Ghosal, Siqi Shen, Navonil Majumder, Rada Mihalcea, and Soujanya Poria. 2022.
\newblock Cicero: A dataset for contextualized commonsense inference in dialogues.
\newblock \emph{arXiv preprint arXiv:2203.13926}.

\bibitem[{He et~al.(2020)He, Fan, Wu, Xie, and Girshick}]{he2020momentum}
Kaiming He, Haoqi Fan, Yuxin Wu, Saining Xie, and Ross Girshick. 2020.
\newblock Momentum contrast for unsupervised visual representation learning.
\newblock In \emph{Proceedings of the IEEE/CVF conference on computer vision and pattern recognition}, pages 9729--9738.

\bibitem[{Hendrycks et~al.(2020)Hendrycks, Burns, Basart, Zou, Mazeika, Song, and Steinhardt}]{hendrycks2020measuring}
Dan Hendrycks, Collin Burns, Steven Basart, Andy Zou, Mantas Mazeika, Dawn Song, and Jacob Steinhardt. 2020.
\newblock Measuring massive multitask language understanding.
\newblock \emph{arXiv preprint arXiv:2009.03300}.

\bibitem[{Jiang et~al.(2023)Jiang, Wu, Lin, Yang, and Qiu}]{jiang2023llmlingua}
Huiqiang Jiang, Qianhui Wu, Chin-Yew Lin, Yuqing Yang, and Lili Qiu. 2023.
\newblock Llmlingua: Compressing prompts for accelerated inference of large language models.
\newblock \emph{arXiv preprint arXiv:2310.05736}.

\bibitem[{Kim et~al.(2022)Kim, Shen, Thorsley, Gholami, Kwon, Hassoun, and Keutzer}]{kim2022learned}
Sehoon Kim, Sheng Shen, David Thorsley, Amir Gholami, Woosuk Kwon, Joseph Hassoun, and Kurt Keutzer. 2022.
\newblock Learned token pruning for transformers.
\newblock In \emph{Proceedings of the 28th ACM SIGKDD Conference on Knowledge Discovery and Data Mining}, pages 784--794.

\bibitem[{Li(2023)}]{li2023unlocking}
Yucheng Li. 2023.
\newblock Unlocking context constraints of llms: Enhancing context efficiency of llms with self-information-based content filtering.
\newblock \emph{arXiv preprint arXiv:2304.12102}.

\bibitem[{Lin(2004)}]{lin2004rouge}
Chin-Yew Lin. 2004.
\newblock Rouge: A package for automatic evaluation of summaries.
\newblock In \emph{Text summarization branches out}, pages 74--81.

\bibitem[{Liu et~al.(2021)Liu, Shen, Zhang, Dolan, Carin, and Chen}]{liu2021makes}
Jiachang Liu, Dinghan Shen, Yizhe Zhang, Bill Dolan, Lawrence Carin, and Weizhu Chen. 2021.
\newblock What makes good in-context examples for gpt-$3 $?
\newblock \emph{arXiv preprint arXiv:2101.06804}.

\bibitem[{Liu et~al.(2024)Liu, Lin, Hewitt, Paranjape, Bevilacqua, Petroni, and Liang}]{liu2024lost}
Nelson~F Liu, Kevin Lin, John Hewitt, Ashwin Paranjape, Michele Bevilacqua, Fabio Petroni, and Percy Liang. 2024.
\newblock Lost in the middle: How language models use long contexts.
\newblock \emph{Transactions of the Association for Computational Linguistics}, 12:157--173.

\bibitem[{Maas et~al.(2011)Maas, Daly, Pham, Huang, Ng, and Potts}]{maas2011learning}
Andrew Maas, Raymond~E Daly, Peter~T Pham, Dan Huang, Andrew~Y Ng, and Christopher Potts. 2011.
\newblock Learning word vectors for sentiment analysis.
\newblock In \emph{Proceedings of the 49th annual meeting of the association for computational linguistics: Human language technologies}, pages 142--150.

\bibitem[{Min et~al.(2022)Min, Lewis, Hajishirzi, and Zettlemoyer}]{min2022noisy}
Sewon Min, Mike Lewis, Hannaneh Hajishirzi, and Luke Zettlemoyer. 2022.
\newblock Noisy channel language model prompting for few-shot text classification.
\newblock In \emph{Proceedings of the 60th Annual Meeting of the Association for Computational Linguistics (Volume 1: Long Papers)}, pages 5316--5330.

\bibitem[{Mu et~al.(2023)Mu, Li, and Goodman}]{mu2023learning}
Jesse Mu, Xiang~Lisa Li, and Noah Goodman. 2023.
\newblock Learning to compress prompts with gist tokens.
\newblock \emph{arXiv preprint arXiv:2304.08467}.

\bibitem[{Narayan et~al.(2018)Narayan, Cohen, and Lapata}]{Narayan2018DontGM}
Shashi Narayan, Shay~B. Cohen, and Mirella Lapata. 2018.
\newblock Don't give me the details, just the summary! topic-aware convolutional neural networks for extreme summarization.
\newblock \emph{ArXiv}, abs/1808.08745.

\bibitem[{Nguyen et~al.(2016)Nguyen, Rosenberg, Song, Gao, Tiwary, Majumder, and Deng}]{nguyen2016ms}
Tri Nguyen, Mir Rosenberg, Xia Song, Jianfeng Gao, Saurabh Tiwary, Rangan Majumder, and Li~Deng. 2016.
\newblock Ms marco: A human generated machine reading comprehension dataset.
\newblock \emph{choice}, 2640:660.

\bibitem[{Qiao et~al.(2024)Qiao, Xie, Gao, Wu, Huang, Fan, Cao, Wang, and Zhang}]{qiao2024dntextspotter}
Qian Qiao, Yu~Xie, Jun Gao, Tianxiang Wu, Shaoyao Huang, Jiaqing Fan, Ziqiang Cao, Zili Wang, and Yue Zhang. 2024.
\newblock Dntextspotter: Arbitrary-shaped scene text spotting via improved denoising training.
\newblock In \emph{Proceedings of the 32nd ACM International Conference on Multimedia}, pages 10134--10143.

\bibitem[{Ram et~al.(2023)Ram, Levine, Dalmedigos, Muhlgay, Shashua, Leyton-Brown, and Shoham}]{ram2023context}
Ori Ram, Yoav Levine, Itay Dalmedigos, Dor Muhlgay, Amnon Shashua, Kevin Leyton-Brown, and Yoav Shoham. 2023.
\newblock In-context retrieval-augmented language models.
\newblock \emph{arXiv preprint arXiv:2302.00083}.

\bibitem[{Reimers and Gurevych(2019)}]{reimers2019sentence}
Nils Reimers and Iryna Gurevych. 2019.
\newblock Sentence-bert: Sentence embeddings using siamese bert-networks.
\newblock \emph{arXiv preprint arXiv:1908.10084}.

\bibitem[{Socher et~al.(2013)Socher, Perelygin, Wu, Chuang, Manning, Ng, and Potts}]{socher2013recursive}
Richard Socher, Alex Perelygin, Jean Wu, Jason Chuang, Christopher~D Manning, Andrew~Y Ng, and Christopher Potts. 2013.
\newblock Recursive deep models for semantic compositionality over a sentiment treebank.
\newblock In \emph{Proceedings of the 2013 conference on empirical methods in natural language processing}, pages 1631--1642.

\bibitem[{Taori et~al.(2023)Taori, Gulrajani, Zhang, Dubois, Li, Guestrin, Liang, and Hashimoto}]{taori2023stanford}
Rohan Taori, Ishaan Gulrajani, Tianyi Zhang, Yann Dubois, Xuechen Li, Carlos Guestrin, Percy Liang, and Tatsunori~B Hashimoto. 2023.
\newblock Stanford alpaca: An instruction-following llama model.

\bibitem[{Team(2023)}]{2023bluelm}
BlueLM Team. 2023.
\newblock Bluelm: An open multilingual 7b language model.
\newblock \url{https://github.com/vivo-ai-lab/BlueLM}.

\bibitem[{Touvron et~al.(2023)Touvron, Lavril, Izacard, Martinet, Lachaux, Lacroix, Rozi{\`e}re, Goyal, Hambro, Azhar et~al.}]{touvron2023llama}
Hugo Touvron, Thibaut Lavril, Gautier Izacard, Xavier Martinet, Marie-Anne Lachaux, Timoth{\'e}e Lacroix, Baptiste Rozi{\`e}re, Naman Goyal, Eric Hambro, Faisal Azhar, et~al. 2023.
\newblock Llama: Open and efficient foundation language models.
\newblock \emph{arXiv preprint arXiv:2302.13971}.

\bibitem[{Wang et~al.(2019)Wang, Singh, Michael, Hill, Levy, and Bowman}]{wang2019glue}
Alex Wang, Amanpreet Singh, Julian Michael, Felix Hill, Omer Levy, and Samuel~R. Bowman. 2019.
\newblock {GLUE}: A multi-task benchmark and analysis platform for natural language understanding.
\newblock In the Proceedings of ICLR.

\bibitem[{Wang et~al.(2023{\natexlab{a}})Wang, Liang, Meng, Shi, Li, Xu, Qu, and Zhou}]{wang12023chatgpt}
Jiaan Wang, Yunlong Liang, Fandong Meng, Haoxiang Shi, Zhixu Li, Jinan Xu, Jianfeng Qu, and Jie Zhou. 2023{\natexlab{a}}.
\newblock Is chatgpt a good nlg evaluator? a preliminary study.
\newblock \emph{arXiv preprint arXiv:2303.04048}.

\bibitem[{Wang et~al.(2023{\natexlab{b}})Wang, Li, Dai, Chen, Zhou, Meng, Zhou, and Sun}]{wang2023label}
Lean Wang, Lei Li, Damai Dai, Deli Chen, Hao Zhou, Fandong Meng, Jie Zhou, and Xu~Sun. 2023{\natexlab{b}}.
\newblock Label words are anchors: An information flow perspective for understanding in-context learning.
\newblock \emph{arXiv preprint arXiv:2305.14160}.

\bibitem[{Wang et~al.(2022{\natexlab{a}})Wang, Yang, Huang, Jiao, Yang, Jiang, Majumder, and Wei}]{wang2022simlm}
Liang Wang, Nan Yang, Xiaolong Huang, Binxing Jiao, Linjun Yang, Daxin Jiang, Rangan Majumder, and Furu Wei. 2022{\natexlab{a}}.
\newblock Simlm: Pre-training with representation bottleneck for dense passage retrieval.
\newblock \emph{arXiv preprint arXiv:2207.02578}.

\bibitem[{Wang et~al.(2024)Wang, Yang, Huang, Yang, Majumder, and Wei}]{wang2024large}
Liang Wang, Nan Yang, Xiaolong Huang, Linjun Yang, Rangan Majumder, and Furu Wei. 2024.
\newblock Large search model: Redefining search stack in the era of llms.
\newblock In \emph{ACM SIGIR Forum}, volume~57, pages 1--16. ACM New York, NY, USA.

\bibitem[{Wang et~al.(2023{\natexlab{c}})Wang, Yang, and Wei}]{wang2023learning}
Liang Wang, Nan Yang, and Furu Wei. 2023{\natexlab{c}}.
\newblock Learning to retrieve in-context examples for large language models.
\newblock \emph{arXiv preprint arXiv:2307.07164}.

\bibitem[{Wang et~al.(2022{\natexlab{b}})Wang, Mishra, Alipoormolabashi, Kordi, Mirzaei, Arunkumar, Ashok, Dhanasekaran, Naik, Stap et~al.}]{wang2022super}
Yizhong Wang, Swaroop Mishra, Pegah Alipoormolabashi, Yeganeh Kordi, Amirreza Mirzaei, Anjana Arunkumar, Arjun Ashok, Arut~Selvan Dhanasekaran, Atharva Naik, David Stap, et~al. 2022{\natexlab{b}}.
\newblock Super-naturalinstructions: Generalization via declarative instructions on 1600+ nlp tasks.
\newblock \emph{arXiv preprint arXiv:2204.07705}.

\bibitem[{Wang et~al.(2023{\natexlab{d}})Wang, Xie, Ding, Feng, and Xia}]{wang2023chatgpt}
Zengzhi Wang, Qiming Xie, Zixiang Ding, Yi~Feng, and Rui Xia. 2023{\natexlab{d}}.
\newblock Is chatgpt a good sentiment analyzer? a preliminary study.
\newblock \emph{arXiv preprint arXiv:2304.04339}.

\bibitem[{Warstadt et~al.(2018)Warstadt, Singh, and Bowman}]{warstadt2018neural}
Alex Warstadt, Amanpreet Singh, and Samuel~R. Bowman. 2018.
\newblock Neural network acceptability judgments.
\newblock \emph{arXiv preprint 1805.12471}.

\bibitem[{Wei et~al.(2023)Wei, Cui, Cheng, Wang, Zhang, Huang, Xie, Xu, Chen, Zhang et~al.}]{wei2023zero}
Xiang Wei, Xingyu Cui, Ning Cheng, Xiaobin Wang, Xin Zhang, Shen Huang, Pengjun Xie, Jinan Xu, Yufeng Chen, Meishan Zhang, et~al. 2023.
\newblock Zero-shot information extraction via chatting with chatgpt.
\newblock \emph{arXiv preprint arXiv:2302.10205}.

\bibitem[{Wingate et~al.(2022)Wingate, Shoeybi, and Sorensen}]{wingate2022prompt}
David Wingate, Mohammad Shoeybi, and Taylor Sorensen. 2022.
\newblock Prompt compression and contrastive conditioning for controllability and toxicity reduction in language models.
\newblock \emph{arXiv preprint arXiv:2210.03162}.

\bibitem[{Wu et~al.(2022)Wu, Rabe, Hutchins, and Szegedy}]{wu2022memorizing}
Yuhuai Wu, Markus~N Rabe, DeLesley Hutchins, and Christian Szegedy. 2022.
\newblock Memorizing transformers.
\newblock \emph{arXiv preprint arXiv:2203.08913}.

\bibitem[{Xie et~al.(2021)Xie, Raghunathan, Liang, and Ma}]{xie2021explanation}
Sang~Michael Xie, Aditi Raghunathan, Percy Liang, and Tengyu Ma. 2021.
\newblock An explanation of in-context learning as implicit bayesian inference.
\newblock \emph{arXiv preprint arXiv:2111.02080}.

\bibitem[{Yang et~al.(2023)Yang, Li, Zhang, Chen, and Cheng}]{yang2023exploring}
Xianjun Yang, Yan Li, Xinlu Zhang, Haifeng Chen, and Wei Cheng. 2023.
\newblock Exploring the limits of chatgpt for query or aspect-based text summarization.
\newblock \emph{arXiv preprint arXiv:2302.08081}.

\bibitem[{Zhang et~al.(2022)Zhang, Roller, Goyal, Artetxe, Chen, Chen, Dewan, Diab, Li, Lin et~al.}]{zhang2022opt}
Susan Zhang, Stephen Roller, Naman Goyal, Mikel Artetxe, Moya Chen, Shuohui Chen, Christopher Dewan, Mona Diab, Xian Li, Xi~Victoria Lin, et~al. 2022.
\newblock Opt: Open pre-trained transformer language models.
\newblock \emph{arXiv preprint arXiv:2205.01068}.

\bibitem[{Zheng et~al.(2023)Zheng, Chiang, Sheng, Zhuang, Wu, Zhuang, Lin, Li, Li, Xing et~al.}]{zheng2023judging}
Lianmin Zheng, Wei-Lin Chiang, Ying Sheng, Siyuan Zhuang, Zhanghao Wu, Yonghao Zhuang, Zi~Lin, Zhuohan Li, Dacheng Li, Eric Xing, et~al. 2023.
\newblock Judging llm-as-a-judge with mt-bench and chatbot arena.
\newblock \emph{arXiv preprint arXiv:2306.05685}.

\bibitem[{Zheng et~al.(2022)Zheng, Wang, and Kong}]{zheng2022linear}
Lin Zheng, Chong Wang, and Lingpeng Kong. 2022.
\newblock Linear complexity randomized self-attention mechanism.
\newblock In \emph{International conference on machine learning}, pages 27011--27041. PMLR.

\end{thebibliography}
\newpage
\clearpage

\appendix
\section{Comparison with Existing Compression Methods}
\label{sec:comparsion}
\begin{table}
\centering
\begin{adjustbox}{width={0.48\textwidth},totalheight={\textheight},keepaspectratio}
\begin{tabular}{l|cccccc}
\bottomrule
\multirow{2}{*}{Methods}  & Additional & Compression  & \# Trainable & Train\\
& Compressor & Tool  & Parameters & Size\\
\hline
LLMLingua~\cite{jiang2023llmlingua} & YES & Pruning &  7B & 57k\\

AutoCompressor~\cite{wingate2022prompt} &NO & Soft Prompt & 7B & UNKNOWN\\

ICAE~\cite{ge2023context} & YES & Soft Prompt  & 70M &240k\\

\hline
UniICL & NO & Soft Prompt & 17M & 47k\\

\toprule
\end{tabular}
\end{adjustbox}
\caption{Comparison among recent compression methods and UniICL. Compression Tool represents the involved compression technique of different methods. Train Size represents the size of the training datasets.}
\label{tab:comparision}
\end{table}

We present a comparison of training costs between UniICL and other recent compression methods in Tab.~\ref{tab:comparision}.

\section{In-Domain Evaluation}
\begin{table}[!h]
\centering
\begin{adjustbox}{width={0.45\textwidth},totalheight={\textheight},keepaspectratio}
\begin{tabular}{llccc|ccc}
\bottomrule
\multirow{2}{*}{Backbone}& \multirow{2}{*}{Method} & \multicolumn{3}{c|}{XSum} & \multicolumn{3}{c}{CICERO} \\
&&R-1&R-2&R-L&R-1&R-2&R-L\\
\hline
\multirow{5}{*}{Vicuna-7b}&Vicuna&19.9&5.0&13.5&17.3&3.3&14.3\\
& \quad\textit{\footnotesize+LoRA} & 25.4 & 7.5 & 17.3 & 28.1 & 10.5 & 25.6\\

&Vicuna-1k&27.3&8.7&19.7&30.5 & 11.3&27.4\\
& \quad\textit{\footnotesize+LoRA}  & 31.2 & 11.0 & 23.1 & 34.1 & 13.5 & 30.2 \\

&UniICL&30.0 &{10.2}&22.3&{32.6}&{12.2}&{28.8}\\
\hline
\multirow{5}{*}{BlueLM-7b}&BlueLM &15.0&3.6&10.4&17.6&3.1&15.0\\
& \quad\textit{\footnotesize+LoRA} & 23.1 & 7.6 & 17.4 & 21.9 & 7.8 & 19.8\\
&BlueLM-1k &28.1& 9.9&22.8&25.1&9.2&23.1\\
& \quad\textit{\footnotesize+LoRA} & 30.8 & 10.5 & 24.6 & 31.2 & 10.8 & 27.4\\
&UniICL&{30.4} &10.2&{23.7}&29.2&10.0&26.6\\
\toprule

\end{tabular}

\end{adjustbox}
\caption{The in-domain results and ablation studies on XSum and CICERO. 1k represents the extended 1k window limitation, while others have a limitation of 512.
}
\label{tab:in-domain}
\end{table}
We conduct the zero-shot in-domain generation evaluation on the entire test set of XSum and CICERO in Tab. \ref{tab:in-domain} by compressing the latter half to virtual tokens and keeping the former unmodified.
UniICL significantly outperforms the baselines, indicating that the compressed virtual tokens can provide the original truncated information by recovering the cut-off parts after supervised fine-tuning. 
Although extending the window to 1k, Vicuna and BlueLM still underperform UniICL, indicating that compressed virtual tokens filter noise information to some extent.

Additionally, to quantify the performance gains brought by the learnable projection layer.
We tune Vicuna and BlueLM with comparable parameters (17M) with LoRA, setting the rank to 32 in Tab.~\ref{tab:in-domain}.
UniICL still outperforms LoRA-adapted LLMs with a 512 window limitation, indicating that the truncation indeed brings performance degradation.

\section{Results on BlueLM}
\label{sec:bluelm}
We also conduct experiments on BlueLM~\cite{2023bluelm} to verify the generality of UniICL.
We demonstrate the result of understanding tasks in Tab.~\ref{tab:understanding-blue}, of the generative tasks in Tab.~\ref{tab:blue-generative}.

\begin{table}[h]
\centering
\begin{adjustbox}{width={\columnwidth},totalheight={\textheight},keepaspectratio}
\begin{tabular}{lc|ccc}
\bottomrule
\multirow{2}{*}{Model} & \multirow{2}{*}{\#-shots}  &CoLA-dev  &SST-2-dev & IMDb \\
&& \multicolumn{3}{c}{Acc.}\\
\hline
\multirow{4}{*}{BlueLM}&0-shot &71.6 & 81.2 & 48.8 \\

&1-shot & 69.6 & 82.6 & 64.8 \\
& 2-shot & 70.0 & 87.0 & 65.6 \\
& 5-shot & 70.5 & 88.6 & 68.7\\
\hline
\multirow{3}{*}{UniICL} &1-shot & 69.6  & 81.2 & 65.4\\
&2-shot & 70.1 & 82.6&  67.0\\
& 5-shot  & 71.8 & 87.0 & 70.4  \\
\hline

\multirow{4}{*}{UniICL$^\spadesuit$}
&1-shot & 70.1 & 80.0 &62.0\\
& 2-shot & 70.3 & 80.8 &67.0 \\
& 5-shot & 71.1 & 85.6 &69.6\\
& 8-shot  & 71.6 & 87.4 & 69.4\\

 \hline
 
\multirow{4}{*}{UniICL$^\spadesuit$ + $L_{ctr}$}
&1-shot & 68.9 & 80.0 & 69.6\\
& 2-shot & 70.6 & 87.2 &  70.6\\
& 5-shot & 71.5 & 89.2 & 71.0 \\
& 8-shot  & 72.4 & 90.4 & 76.8\\

\toprule
\end{tabular}
\end{adjustbox}
\caption{The ICL results of understanding tasks with the backbone of BlueLM.}
\label{tab:understanding-blue}
\end{table}

\begin{table}[h]
\centering
\begin{adjustbox}{width={\columnwidth},totalheight={\textheight},keepaspectratio}
\begin{tabular}{lc|ccc|ccc}
\bottomrule
\multirow{2}{*}{Method} & \multirow{2}{*}{\#-shots}  &\multicolumn{3}{c|}{XSum} &\multicolumn{3}{c}{Arxiv} \\
&&R-1&R-2&R-L &R-1&R-2&R-L\\
\hline
\multirow{2}{*}{BlueLM} & 0-shot &15.0&3.6&10.4 &30.9& 7.7&24.7\\
 & 1-shot &19.1& 4.8&12.1&23.0& 3.6&19.0\\
\hline
\multirow{3}{*}{UniICL} & 1-shot &24.0 &6.9  &18.0 &31.4 &7.7 & 25.2\\
 & 2-shot &25.0  &7.3   &18.8&30.8	&7.3 &24.8\\
 & 5-shot &25.3 &7.4 &19.1 &31.9 & 7.8 & 26.0\\

\hline

\multirow{3}{*}{UniICL$^\spadesuit$} &1-shot &24.7 & 7.2 &18.5 &31.0& 7.5& 24.9\\
&2-shot &25.1 &  7.4& 19.0& 31.2& 7.7& 25.1 \\

&5-shot &26.3 &7.6& 20.0& 31.5& 7.9& 25.3\\

\hline
\multirow{3}{*}{UniICL$^\spadesuit$ + $L_{ctr}$ } &1-shot &25.2 &7.4&18.9&31.6&7.9&25.4\\
 &2-shot &{25.4}&{7.6}&{19.1}&{31.9}&{8.0}&{25.6}\\

  &5-shot &{26.5}&{7.9}&{20.3}&{32.1}&{8.0}&{25.5}\\

\toprule
\end{tabular}
\end{adjustbox}
\caption{The ICL results of generative tasks with the backbone of BlueLM.}
\label{tab:blue-generative}
\end{table}

\section{Supplementary Ablation on Llama2}
\label{sec:llama2}
AutoCompressor~\cite{wingate2022prompt} and ICAE~\cite{ge2023context} are built on Llama2-7B~\cite{touvron2023llama}, which are soft-prompt-based methods similar to UniICL.
Therefore, we evaluate UniICL with Llama2 as the backbone.
As shown in Tab~\ref{tab:llama2-understanding} and Tab.~\ref{tab:llama2-generative}, UniICL achieves substantial improvements compared with unmodified Llama2 and outperforms ICAE and AutoCompressor demonstrated in Tab.~\ref{tab:understanding}.

\begin{table}[ht]
\centering
\begin{adjustbox}{width={\columnwidth},totalheight={\textheight},keepaspectratio}
\begin{tabular}{lc|ccc}
\bottomrule
\multirow{2}{*}{Model} & \multirow{2}{*}{\#-shots}  &CoLA-dev  &SST-2-dev & IMDb \\
&& \multicolumn{3}{c}{Acc.}\\
\hline
\multirow{4}{*}{Llama2}
&0-shot    & 73.4 & 93.0 & 85.3 \\
&1-shot    & 74.8 & 94.0 & 85.5 \\
&2-shot    & 75.6 & 94.9 & 87.8 \\
&5-shot    & 84.3 & 97.2 & 92.7 \\
\hline

\multirow{3}{*}{UniICL} 
&1-shot    & 74.9   & 94.1   & 94.6   \\
&2-shot    & 75.9   & 95.1   & 96.1   \\
&5-shot    & 85.4   & 95.7   & 96.5   \\

\toprule
\end{tabular}
\end{adjustbox}
\caption{The ICL results of understanding tasks with the backbone of Llama2.}
\label{tab:llama2-understanding}
\end{table}

\begin{table}[h]
\centering
\begin{adjustbox}{width={\columnwidth},totalheight={\textheight},keepaspectratio}
\begin{tabular}{lc|ccc|ccc}
\bottomrule
\multirow{2}{*}{Method} & \multirow{2}{*}{\#-shots}  &\multicolumn{3}{c|}{XSum} &\multicolumn{3}{c}{Arxiv} \\
&&R-1&R-2&R-L &R-1&R-2&R-L\\
\hline
\multirow{2}{*}{Llama2} 
 & 0-shot &27.4 &7.6 &20.1 &32.9& 8.9&29.2\\
 & 1-shot &27.7& 7.9 &20.3 &30.1& 8.0 &28.4\\
\hline

\multirow{3}{*}{UniICL} 
&1-shot &27.8 & 8.0 &20.5 &33.3 & 9.2 & 29.7\\

&2-shot &28.4 & 8.6 & 21.3 & 34.0 & 9.4	& 30.3\\

&5-shot &29.3 &9.1& 22.0 & 34.5& 9.7 & 30.8\\
\toprule
\end{tabular}
\end{adjustbox}
\caption{The ICL results of generative tasks with the backbone of LLama2.}
\label{tab:llama2-generative}
\end{table}

\section{Compression Ratio Selection on Different Tasks}
\label{sec:ratioselection}
\begin{figure}
    \centering
    \includegraphics[width=\linewidth]{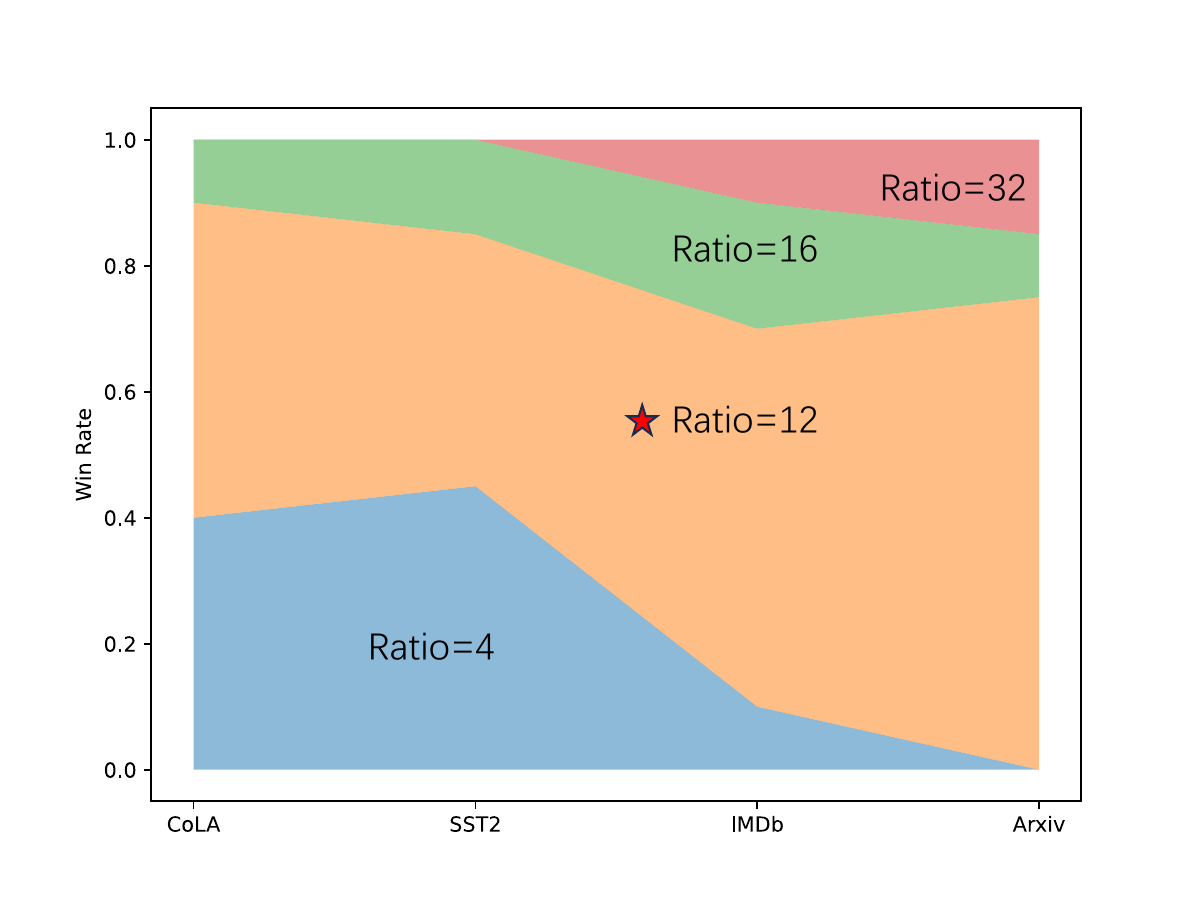}
    \caption{Winrate of different compression ratios on out-of-domain evaluation in 1-shot settings.}
    \label{fig:winrate}
\end{figure}
We illustrate suitable ratio selection across four out-of-domain datasets in Fig.~\ref{fig:winrate}.
For tasks with relatively short inputs, such as CoLA and SST2, UniICL tends to perform better with a compression ratio set to 4.
While in IMDb and Arxiv, which are longer, UniICL performs better with higher compression ratios.
UniICL with a 12$\times$ compression ratio substantially outperforms other settings on four datasets.
\begin{figure}[t]
    \centering
    \includegraphics[width=\linewidth]{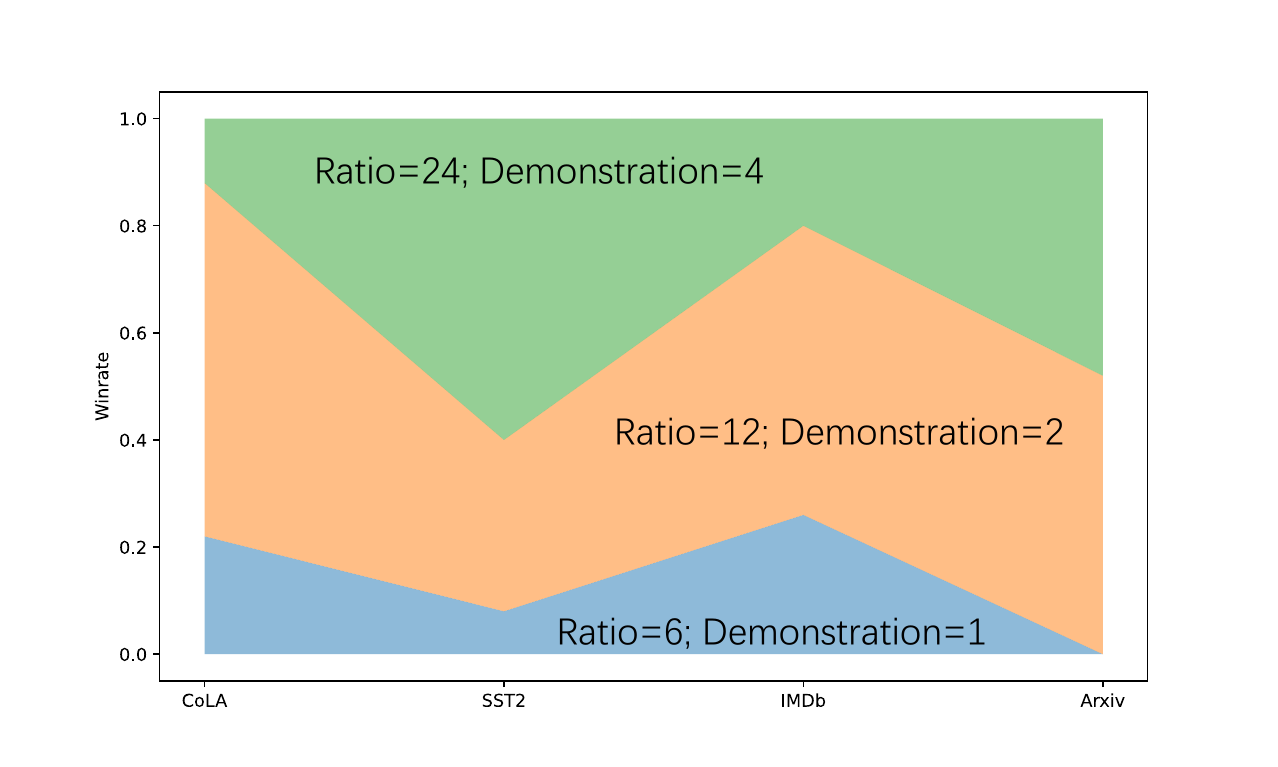}
    \caption{Winrate of UniICL with a fixed number of Memory Tokens.}
    \label{fig:winrate_2}
\end{figure}
Additionally, we are curious about whether it is necessary to introduce more demonstrations with a higher compression ratio.
In Fig.~\ref{fig:winrate_2}, we find that the performance of compressing 2 demonstrations with a $12\times$ ratio is stable and outperforms other settings across 3/4 datasets.
6$\times$ compression ratio with 1 demonstration compressed performs worst in general.
When compressing 4 demonstrations with a $24\times$ ratio, its performance is comparable, and it slightly outperforms the $12\times$ ratio in SST2.

\section{Fixed Compression Ratio Training}
\label{sec:fixedratio}
\begin{figure}
    \centering
    \includegraphics[width=\linewidth]{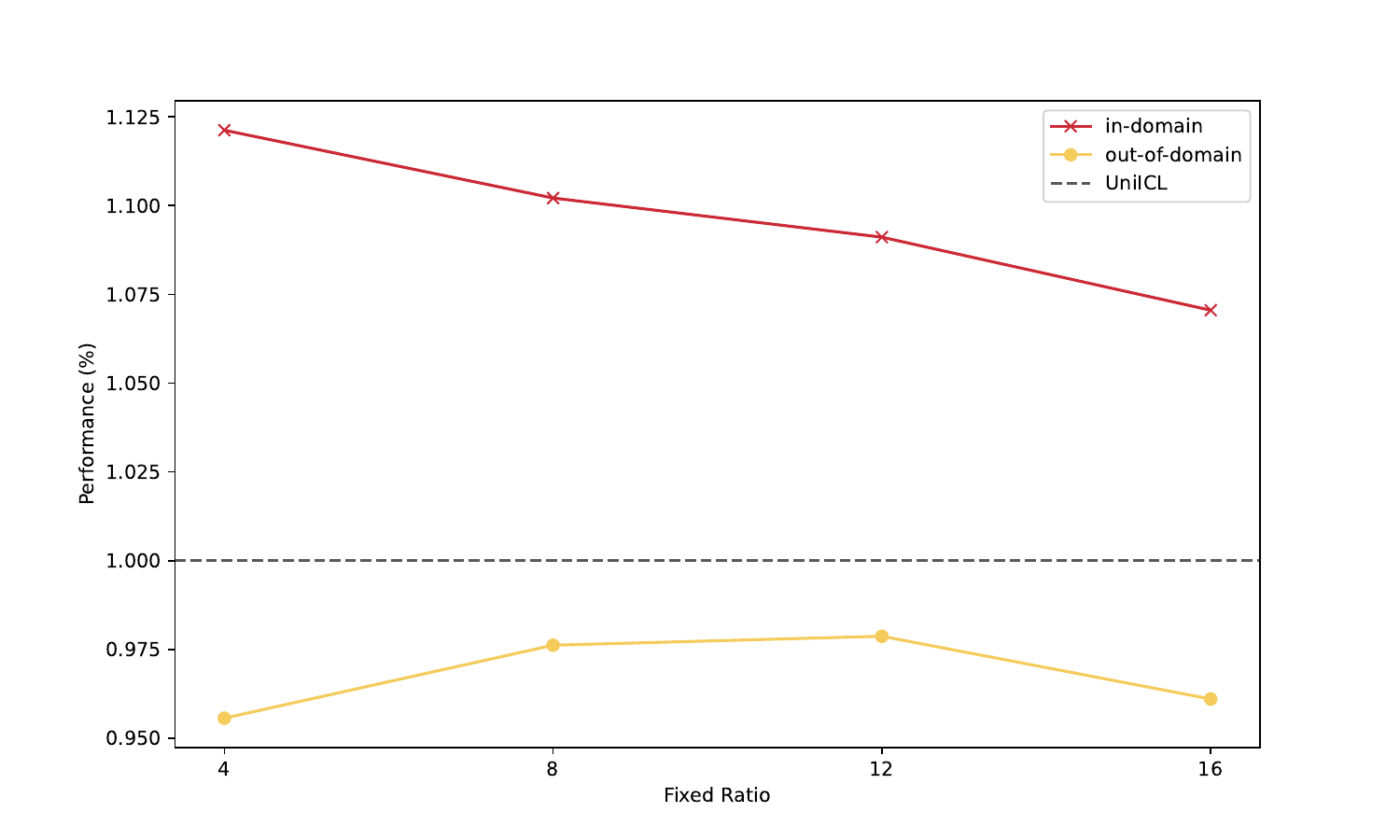}
    \caption{The relative performance on in-domain and out-of-domain datasets, with UniICL trained with a fixed ratio. Out-of-domain evaluation applies 1-shot settings.}
    \label{fig:fixratio}
\end{figure}
To verify the effectiveness of the dynamic sampled compression ratio of UniICL, we train models with more extensive fixed compression ratios and perform out-of-domain evaluation with the same ratio in Fig.~\ref{fig:fixratio}.
Results indicate that fixed compression ratios work better than dynamic sampled ratios in in-domain evaluation, but underperform in out-of-domain evaluation.
We attribute this to the fixed compression ratio makes models exhibit over-fitting during training, and demonstration compression degrades to Prefix Tuning.

\section{Visualization of Memory Tokens}
\label{sec:visualization}
\begin{figure*}[t]
    \centering
    \includegraphics[width=0.85\textwidth]{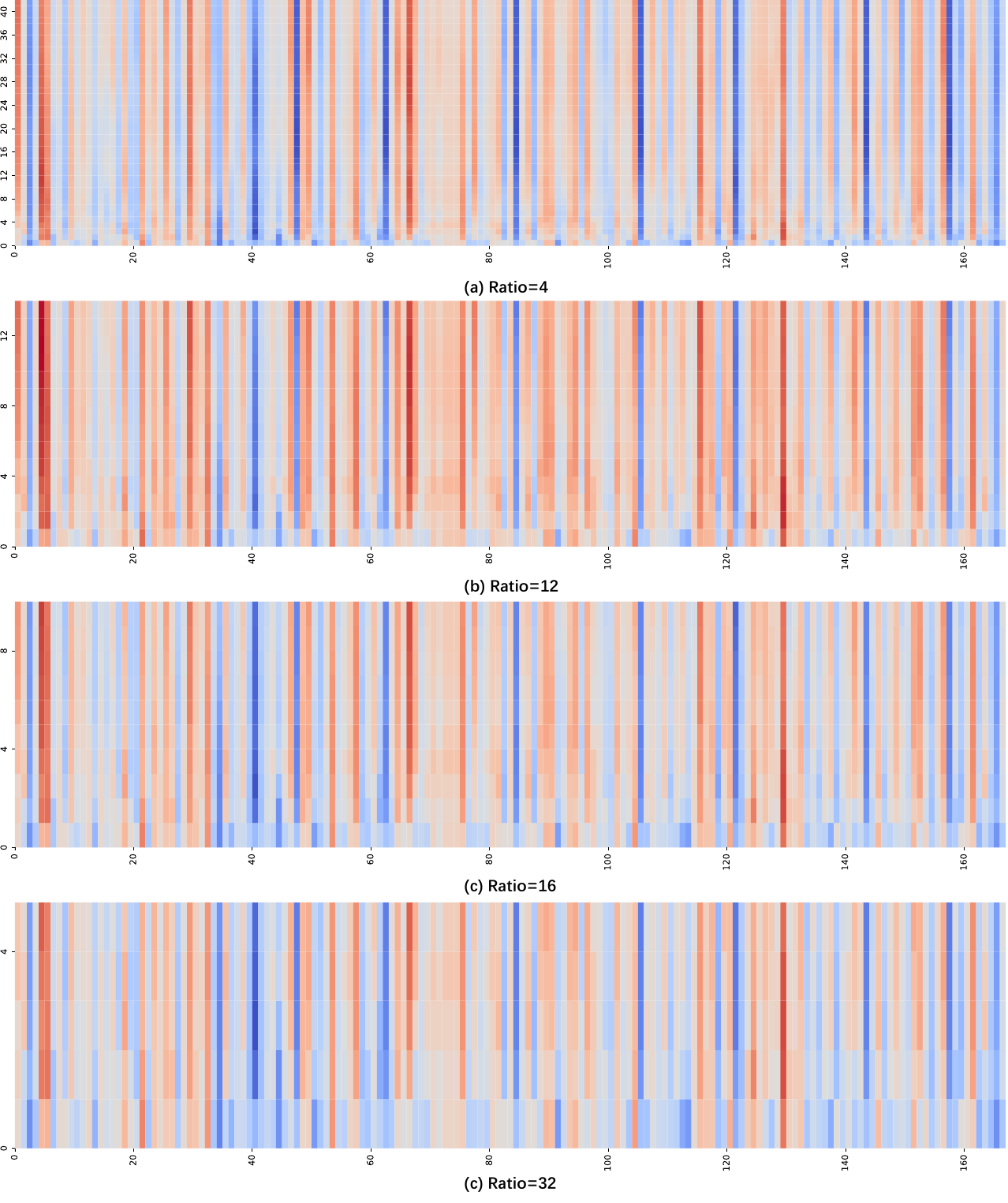}
    \caption{Cosine similarity between Memory Tokens (vertical axis) and original embeddings (horizon axis). }
    \label{fig:sims}
\end{figure*}

\begin{figure}[t]
    \centering
    \includegraphics[width=0.75\columnwidth]{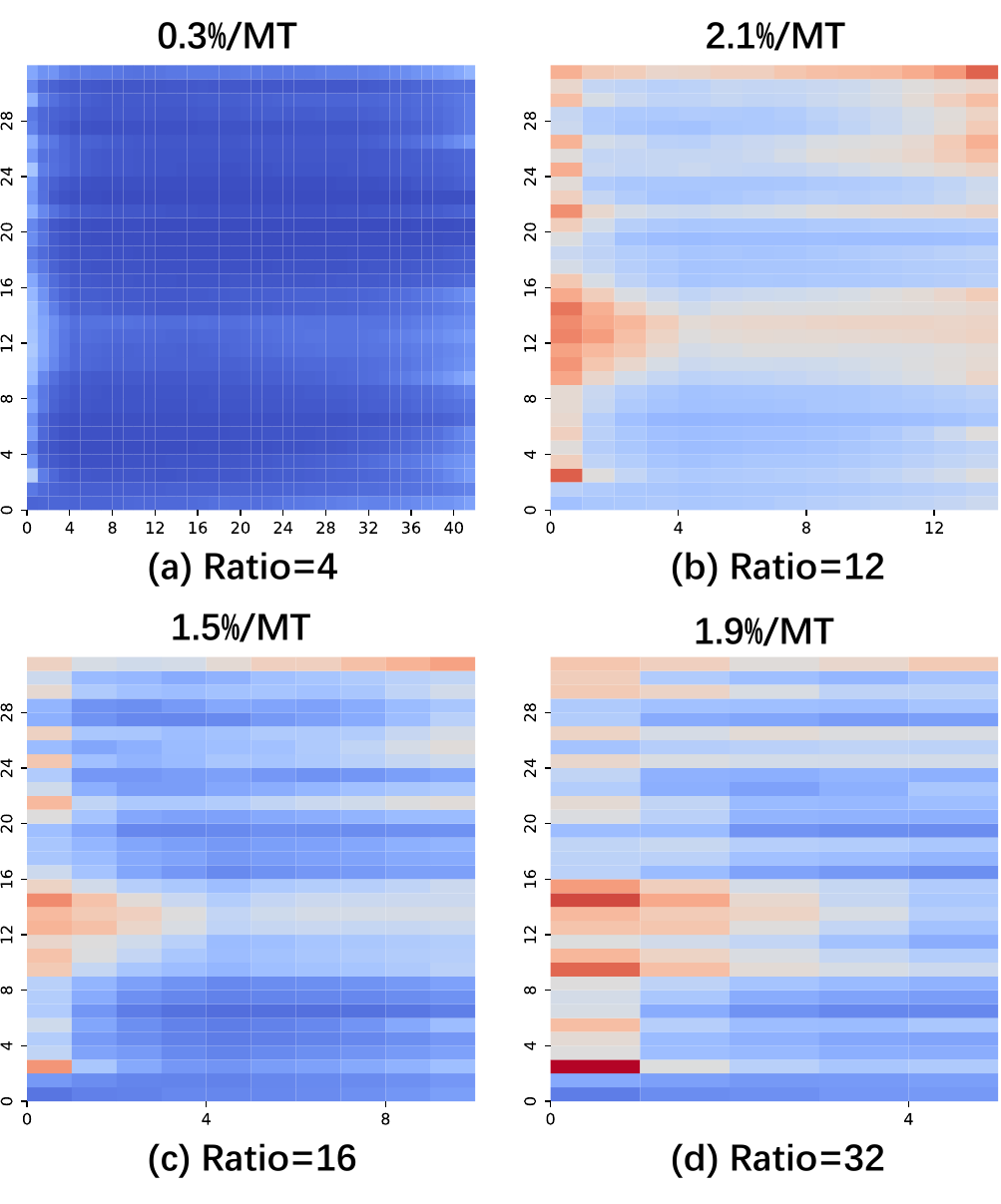}
    \caption{Attention scores on Memory Tokens in the first step generation. The vertical axis describes the 32 LLM layer, and the horizon axis indicates the number of Memory Tokens across different compression ratios. Above each figure, \textbf{\%/MT} represents the average proportion of the attention score occupied by memory tokens in each LLM layer.}
    \label{fig:Attn}
\end{figure}

To explore how Memory Tokens work within UniICL across different compression ratios, we visualize the cosine similarity between Memory Tokens and original embeddings in Fig.~\ref{fig:sims} and attention scores of the first generation step in Fig.~\ref{fig:Attn}.

Intuitively, the 4$\times$ compression ratio should retain more information due to more Memory Tokens.
However, as shown in Fig.~\ref{fig:sims}(a), the cosine similarity is relatively sparser than the 4$\times$ compression ratio illustrated in Fig.~\ref{fig:sims}(b).
This tendency is aligned with the first step attention scores in Fig.~\ref{fig:Attn}(a).
According to merely 0.3\% average attention occupied in a generation, we can conclude that more Memory Tokens fail to provide models with more information.
We attribute this phenomenon to the given semantic information being distributed over all Memory Tokens as models attend to each Memory Token equally in Fig.~\ref{fig:Attn}(a).
Fewer Memory Tokens are enough to concentrate this information, represented as relatively concentrated similarity distribution in Fig.~\ref{fig:sims}(b) and higher attention scores in Fig.~\ref{fig:Attn}(b), both of which indicate denser information retained.
When the compression ratio becomes higher, such as 16 or 32, Memory Tokens become fewer and therefore sparse information retrained as shown in Fig.~\ref{fig:sims}(c), Fig.~\ref{fig:sims}(c), Fig.~\ref{fig:Attn}(c), and Fig.~\ref{fig:Attn}(c).
This also provides an explanation for the slow performance degradation with ratios varying from 4 to 12 and drops sharply at 16 in Fig.~\ref{fig:ratio}.

\section{Datasets \& Metrics}
\label{sec:DM}
\paragraph{Datasets}
We mix up three public datasets for compression and selection augmentation training, described in Tab. \ref{tab:dataset}.
The training set includes an instruction dataset, SUPER-NI, which we used to make UniICL respond to various instructions.
Notably, we don't perform an in-domain evaluation on SUPER-NI as it only contains a training set.
After training, we extensively evaluate UniICL on out-of-domain evaluation, involving text summarization~\cite{Narayan2018DontGM}, passage ranking~\cite{nguyen2016ms}, sentiment classification~\cite{maas2011learning,socher2013recursive}, linguistic acceptability~\cite{warstadt2018neural}, and a popular reasoning benchmark~\cite{hendrycks2020measuring}, more details referring to Tab. \ref{tab:settings}.
MS MARCO is popularly used in Information Retrieval (IR), we use this dataset to evaluate the ability of UniICL to capture document-level information.
Specifically, MMLU (Massive Multitask Language Understanding) is a comprehensive evaluation benchmark including 57 subjects from STEM, Humanities, Social Sciences, and Other fields.
We use this benchmark to evaluate the reasoning ability of UniICL.
UniICL selects demonstrations from its training set in high-resource ICL, and we fixed the number of candidate demonstrations to 20 for low-resource ICL evaluation.

\paragraph{Evaluation Metrics}
ROUGE \cite{lin2004rouge} is a widely adopted metric in many generative tasks that evaluates how similar the generated hypothesis is to the golden label.
Therefore, ROUGE is used in our experiments to evaluate the quality responses generated conditioned on compressed virtual tokens, and we report the F-1 scores of ROUGE-1, ROUGE-2, and ROUGE-L (abbreviated R-1, R-2, R-L in the following), and we employed the files2rouge \footnote{\url{https://github.com/pltrdy/files2rouge.}} library in practice.
Following the previous works, we report the accuracy of close-ended evaluation and MRR@10 for passage ranking.

\end{document}